%% file: main.tex
\documentclass[conference]{IEEEtran}
\usepackage{blindtext, graphicx}
\ifCLASSINFOpdf
\else
\fi
\usepackage{url}

\usepackage{todonotes}
\usepackage{algorithm}
\usepackage{algorithmic}
\usepackage{amsmath}
\usepackage{amssymb}
\usepackage{hyperref}
\usepackage{float}

\input{commands}

\begin{document}
%
\title{Trajectory Optimization Through Contacts and Automatic Gait Discovery for Quadrupeds}

\author{\IEEEauthorblockN{Michael Neunert, Farbod Farshidian, Alexander W. Winkler, Jonas Buchli}
\IEEEauthorblockA{Agile \& Dexterous Robotics Lab\\
ETH Z\"urich, Switzerland\\
Email: \{neunertm, farbodf, winklera, buchlij\} @ethz.ch}
}


%


\maketitle

\begin{abstract}
In this work we present a trajectory Optimization framework for whole-body motion planning through contacts. We demonstrate how the proposed approach can be applied to automatically discover different gaits and dynamic motions on a quadruped robot. In contrast to most previous methods, we do not pre-specify contact switches, timings, points or gait patterns, but they are a direct outcome of the optimization. Furthermore, we optimize over the entire dynamics of the robot, which enables the optimizer to fully leverage the capabilities of the robot. To illustrate the spectrum of achievable motions, here we show eight different tasks, which would require very different control structures when solved with state-of-the-art methods. Using our trajectory Optimization approach, we are solving each task with a simple, high level cost function and without any changes in the control structure. Furthermore, we fully integrated our approach with the robot's control and estimation framework such that optimization can be run online. By demonstrating a rough manipulation task with multiple dynamic contact switches, we exemplarily show how optimized trajectories and control inputs can be directly applied to hardware. 
\end{abstract}


%
\IEEEpeerreviewmaketitle

\section{Introduction}
In robotics motion planning and control, one major challenge is to specify a high level task for a robot without specifying how to solve this task. For legged locomotion such a task include a goal position to reach or to manipulate an object without specifying gaits, contacts, balancing or other behaviors. Trajectory Optimization recently gained a lot of attention in robotics research since it promises to solve some of these problems. It could potentially solve complex motion planning tasks for robots with many degrees of freedom, leveraging the full dynamics of the system. Yet, there are two challenges of optimization optimization. Firstly, Trajectory Optimization problems are hard problems to solve, especially for robots with many degrees of freedom and for robots that make or break contact. Therefore, many approaches add heuristics or pre-specify contact points or sequences. However, this then defines again how the robot is supposed to solve the task, affecting optimality and generality. Secondly, Trajectory Optimization cannot be blindly applied to hardware but requires an accurate model as well as a good control and estimation framework. In this work, we are addressing both issues. In our Trajectory Optimization framework, we only specify high level tasks, allowing the solver to find the optimal solution to the problem, optimizing over the entire dynamics and automatically discovering the contact sequences and timings. Second, we also demonstrate how such a dynamic task can be generated online. This work does not present a general tracking controller suitable for all trajectories. However, we show the successful integration of our Trajectory Optimization with our estimation and control framework allowing for executing a selection of tasks even under disturbances.

\begin{figure}[htbp]
\centering
\includegraphics[width=\columnwidth]{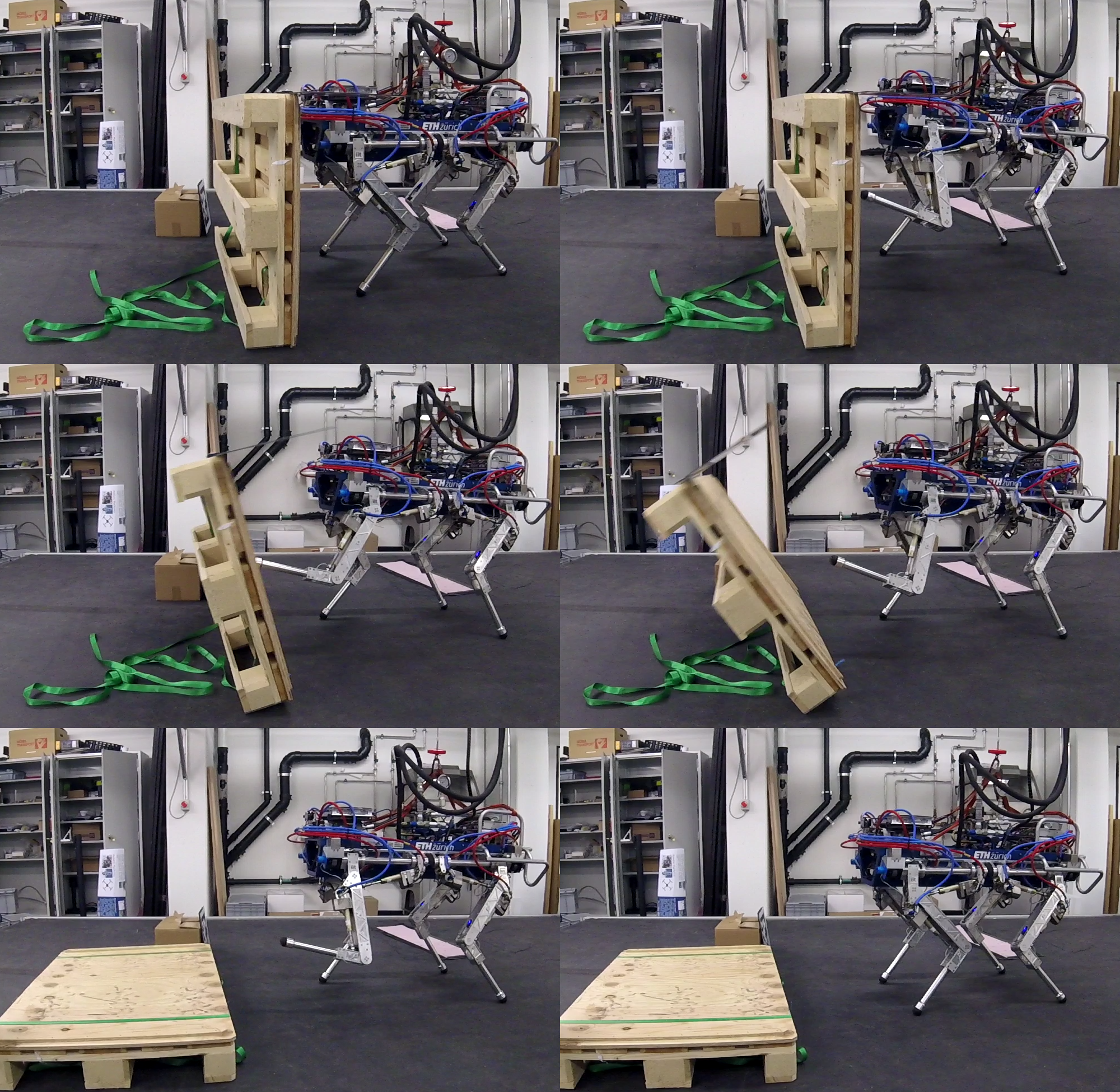}
\caption{Sequence of images during execution of the rough manipulation task in hardware. The time series starts at the top left corner and time progresses line-wise. The motion shows how the shift of the main body, lift-off, push-over and return to default stance.}
\label{fig:hw_sequence}
\end{figure}

Trajectory Optimization tries to solve a general, time-varying, non-linear optimal control problem. There are various forms of defining and solving the control problem. An overview can be found in \cite{betts1998survey}. With the increase of computational power, Trajectory Optimization can be applied to higher dimensional systems like legged robots. Thus, it has gained a lot of attention in recent years and impressive results have been demonstrated in simulation \cite{posa2014direct,mordatch2012discovery,tassa2012synthesis}. One of the conceptually closest related work is \cite{tassa2014control}. However, no gait discovery or very dynamic motions are shown and no hardware results are presented. Later work \cite{koenemann2015whole} includes hardware results but the planning horizon is short, the motions are quasi-static and contact changes are slow. In \cite{mastalli} Trajectory Optimization through contacts is demonstrated on hardware. While the results are promising, the approach is only tested on a low-dimensional, single leg platform and for very simple tasks. In general, work that demonstrates Trajectory Optimization through contacts applied to physical, legged systems is very rare. Automatically discovered gaits and dynamic motions demonstrated on quadruped hardware are presented in \cite{gehring2016}. The presented results are very convincing and they are also considering actuator dynamics. However, their approach differs in key areas to the presented work: Contact sequences are pre-specified and they optimize over control parameters for a fixed control structure rather than whole body trajectories and control inputs. Additionally, they are using black box optimization. However, the tracking controllers employed in both approaches are very similar.

\subsection{Contributions}
In this work, we apply Trajectory Optimization (TO) in form of Sequential Liner-Quadratic (SLQ) control \cite{slq} to the quadruped robot HyQ \cite{hyq}, both in simulation and on hardware. Our work is one of very few examples, where TO is used to plan dynamic whole body motions through contact changes and where some of these trajectories are applied to hardware. Furthermore, this is possibly the first time that a whole body TO approach is shown on quadruped hardware. In contrast to state-of-the-art approaches, neither contact switching times nor contact events nor contact points are defined a priory. Instead they are a direct outcome of the optimization. This allows us to generate a diverse set of motions and gaits using a single approach. By using an efficient formulation based on Differential Dynamic Programming and a customized, high-performance solver, optimization times for these tasks usually do not exceed a minute, even for complex trajectories. Thus, while the Trajectory Optimization is not performed in a control loop, it is still run online. Hence, this work is also one of the earliest examples of whole body TO fully integrated with a control, state estimation and ground estimation pipeline of a legged robotic system.

\section{Trajectory Optimization}
\subsection{Optimal Control Problem}
In this work, we consider a general non-linear system of the following form
\begin{align}
\dot{\vx}(t) = f(\vx(t),\vu(t))
\label{eq:system_model}
\end{align}
where $x(t)$ and $u(t)$ denote state and input trajectories respectively. In Trajectory Optimization, we indirectly optimize these trajectories by altering a linear, time-varying feedback and feedforward controller of the form
\begin{align}
\vu(\vx,t) = \vu_{ff}(t) + \mathcal{K}(t) \vx(t) 
\label{control_law}
\end{align}
where $\mathcal{K}(t)$ is a time-varying control gain matrix and $u_{ff}(t)$ a time-varying feedforward control action. Our Trajectory Optimization approach then tries to solve a finite-horizon optimal control problem which minimizes a given cost function of the following form
\begin{align}
J(\vx,\vu) = h\left( \vx(t_f) \right) + \int_{t=0}^{t_f}l \left(  
\vx(t),\vu(t)
\right) dt
\label{eq:cost_function}
\end{align}
\subsection{Sequential Linear Quadratic Control}
Sequential Linear Quadratic Control is an iterative optimal-control algorithm. SLQ first rolls out the system dynamics. Then, the non-linear system dynamics are linearized around the trajectory and a quadratic approximation of the cost function is computed. The resulting Linear-Quadratic Optimal Control problem is solved backwards. Since the solution to the Linear-Quadratic Optimal Control problem can be computed in closed form with Ricatti-like equations, SLQ is very efficient. Algorithm \ref{alg:slq} summarizes the algorithm. SLQ computes both, a feedforward control action as well as a time-varying linear-quadratic regulator that stabilizes the trajectory.

\begin{algorithm}[tpb] \caption{SLQ Algorithm} \label{alg:slq}
\begin{algorithmic} \scriptsize \STATE \textbf{Given} 
\STATE - System dynamics:
$\vx(t+1)=\vf\left(\vx(t),\vu(t)\right)$ 
\STATE - Cost function $J = h\left( \vx(t_f) \right) + \sum\limits_{t=0}^{t_f-1}{l \left( \vx(t),\vu(t) \right)}$ 
\STATE - Initial stable control law, $\mathbf{\vu}(\vx,t)$ 
\REPEAT \STATE - t(0) simulate the system dynamics: 
\STATE $\tau: \vx_n(0),\vu_n(0),\vx_n(1),\vu_n(1),\dots,\vx_n(t_f-1),\vu_n(t_f-1),\vx_n(t_f)$
\STATE - Linearize the system dynamics along the trajectory $\tau$: 
\STATE
$\vdx(t+1) = \vA(t)\vdx(t)+\vB(t)\vdu(t)$ 
\STATE \qquad $\{ \vA(t) \}_{0}^{t_f-1}$, $\vA(t)=\frac{\partial\vf}{\partial \vx}|_{\vx(t),\vu(t)}$ 
\STATE \qquad $\{ \vB(t) \}_{0}^{t_f-1}$, $\vB(t)=\frac{\partial\vf}{\partial \vu}|_{\vx(t),\vu(t)}$
\STATE - Quadratize cost function along the trajectory $\tau$: 
\STATE $\tilde{J}
\approx p(t)+\vdx^T(t_f)\vp(t_f)+\frac{1}{2}\vdx^T(t_f)\vP(t_f)\vdx(t_f) $
\STATE \quad \hspace{2ex} $+ \sum\limits_{t=0}^{t_f-1}q(t)+\vdx^T\vq(t)+\vdu^T\vr(t) $
\STATE \quad \hspace{2ex} $+ \frac{1}{2}\vdx^T\vQ(t)\vdx+\frac{1}{2}\vdu^T\vR(t)\vdu$ 
\STATE - Backwards solve the Riccati-like difference equations: 
\STATE \qquad $\vP(t) = \vQ(t)+\vA^T(t)\vP(t+1)\vA(t)+ $
\STATE \hspace{12ex} $\vK^T(t)\vH\vK(t) +  \vK^T(t)\vG+\vG^T\vK(t)$ 
\STATE \qquad $\vp(t) = \vq+\vA^T(t)\vp(t+1)+\vK^T(t)\vH\vl(t)+\vK^T(t)\vg+\vG^T\vl(t)$
\STATE \qquad $\vH = \vR(t)+\vB^T(t)\vP(t+1)\vB(t)$ 
\STATE \qquad $\vG = \vB^T(t)\vP(t+1)\vA(t)$ 
\STATE \qquad $\vg = \vr(t)+\vB^T(t)\vp(t+1)$ 
\STATE \qquad $\vK(t)= -\vH^{-1}\vG$ \% feedback update 
\STATE \qquad $\vl(t)= -\vH^{-1}\vg$ \% feedforward increment 
\STATE \textbf{Line search}
\STATE - Initialize $\alpha = 1$
\REPEAT 
\STATE - Update the control: 
\STATE $\mathbf{\vu}(\vx,t) = \vu_n(t) + \alpha \vl(t) + \vK(t)
\left(\vx(t)-\vx_n(t)\right)$ 
\STATE - Forward simulate the system dynamics: 
\STATE $\tau:
\vx_n(0),\vu_n(0),\vx_n(1),\dots,\vx_n(t_f-1),\vu_n(t_f-1),\vx_n(t_f)$
\STATE - Compute new cost: 
\STATE $J = h\left(\vx(t_f) \right) + \sum_{t=0}^{t_f-1}{l \left( \vx(t),\vu(t) \right) dt}$
\STATE - decrease $\alpha$ by a constant $\alpha_{d}$:
\STATE $\alpha = \alpha / \alpha_{d}$
\UNTIL{found lower cost or number of maximum line search steps reached}
\UNTIL{maximum number of iterations or converged ($\vl(t) < l_t$)}
\end{algorithmic} \end{algorithm}

One drawback of SLQ is that it cannot handle hard state constraints. While one can add state constraints as soft constraints into the cost function, they can result in bad numerical behavior. Handling input constraints in these type of algorithms is possible, as shown in \cite{ilqg}, we are not considering them as of now. Even though we are not considering constraints in this work, most tasks result in constraint satisfactory trajectories within the joint position and torque limits.

\subsection{Cost Function}
While SLQ can handle non-quadratic cost functions, pure quadratic cost functions increase convergence and are often sufficient to obtain the desired behaviour. Therefore, in this work we assume our cost function to be of quadratic form
\begin{align}
J &= \bar{\vx}(t_f)^T \vH \bar{\vx}(t_f) + \int\limits_{t=0}^{t_f}{
\bar{\vx}(t)^T \vQ \bar{\vx}(t) + \bar{\vu}(t)^T \vR \bar{\vu}(t)
} \nonumber \\ &+ W(\vx,t) \, dt
\label{eq:cost_function_experiments}
\end{align}
where $\bar{x}(t)$ and $\bar{u}(t)$ represent deviations of state and input from a desired state and input respectively. $H$, $Q$ and $R$ are the weightings for final cost, intermediate cost and input cost respectively. Additionally, we add intermediate state waypoints to guide the trajectory. These are weighted by the waypoint cost matrix $W(x,t)$ defined as
\begin{align}
W(\vx,t) =  \sum\limits_{n=0}^{N}{
\hat{\vx}(t)^T \vW_p \hat{\vx}(t)
\sqrt{\frac{\rho_p}{2 \pi}} \exp{\left( - \frac{\rho_p}{2} (t - t_p)^2 \right)}
}
\label{eq:waypoint}
\end{align}

\section{System Model and Robot Description}
For the experiments in this paper, we use the hydraulically actuated, quadrupedal robot HyQ \cite{hyq}. Each of the four legs on HyQ has three degrees of freedom, namely hip abduction/adduction (HAA), hip flexion/extension (HFE) and knee flexion/extension (KFE). Each of these joints is driven by a hydraulic actuator which is controlled by a hydraulic valve. The joint torque and position is measured via load cells and encoders respectively. A hydraulic force control loop is closed at joint level, providing joint torque reference tracking. 

\subsection{Rigid Body Dynamics}
While it is a simplification, torque tracking performance on HyQ is sufficient for modeling it as a perfectly torque controlled robot. Thus, we assume HyQ to behave like a rigid body system defined as

\begin{equation}
    \vM(\vq)\ddot{\vq} + \vC(\vq,\dot{\vq}) + \vG(\vq) = \vJ^T_c \lambda(\vq, \dot{\vq}) + \vS^T \tau
    \label{eq:rbd}
\end{equation}

where $M$ denotes the inertia matrix, $C$ the centripetal and coriolis forces and $G$ gravity terms. $q$ is the state vector containing the 6 DoF base state as well as joint positions and velocities. External and contact forces $\lambda$ act on the system via the contact Jacobian $J_c$. Torques created by the actuation system $\tau$ are mapped to the states via the selection matrix $S$.

To formulate our dynamics according to Equation \ref{eq:system_model}, we define our state as
\begin{equation}
    \vx = [{}_{\mathcal{W}}\vq~{}_{\mathcal{L}}\dot{\vq}]^T = [{}_{\mathcal{W}}\vq_B~{}_{\mathcal{W}}\vx_B~\vq_J~{}_{\mathcal{L}}\omega_B~{}_{\mathcal{L}}\dot{\vx}_B~\dot{\vq}_J]^T
    \label{eq:coordinates}
\end{equation}
where ${}_{\mathcal{W}}\vq_B$ and ${}_{\mathcal{W}}\vx_B$ define base orientation and position respectively, which are expressed in a global inertial ``world'' frame $\mathcal{W}$. The base orientation is expressed in euler angles (roll-pitch-yaw). The base's angular and linear velocity are denoted as $\omega_B$ and $\dot{\vx_B}$ respectively and both quantities are expressed in a local body frame $\mathcal{L}$. Joint angles and velocities are expressed as $\vq_J$ and $\dot{\vq}_J$ respectively. Expressing base pose and twist in different frames allows for more intuitive tuning of the cost function weights. Using Equations \ref{eq:rbd} and \ref{eq:coordinates} our system dynamics in Equation \ref{eq:system_model} become
\begin{align}
\dot{\vx}(t) &=
             \begin{bmatrix}
                {}_{\mathcal{W}}\dot{\vq} \\ 
                {}_{\mathcal{L}}\ddot{\vq}
             \end{bmatrix}
             \\ &= 
             \begin{bmatrix}
                \vR_{\mathcal{W}\mathcal{L}}~{}_{\mathcal{L}}\dot{\vq} \\
                \vM^{-1}(\vq)(\vS^T\tau + \vJ_c \lambda(\vq,\dot{\vq}) - \vC(\vq, \dot{\vq}) - \vG(\vq))
             \end{bmatrix} \nonumber
\label{eq:system_model}
\end{align}
where $\vR_{\mathcal{W}\mathcal{L}}$ defines the rotation between the local body frame $\mathcal{L}$ and the inertial ``world'' frame $\mathcal{W}$. While dropped in our notation for readability, $\vR_{\mathcal{W}\mathcal{L}}$ is a function of ${}_{\mathcal{W}}\vq$ which needs to be considered during linearization.

For SLQ we need to linearize the system given in in Equation \ref{eq:system_model}. For the derivatives of the upper row with respect to the state $\vx$ as well as all derivatives with respect to $\tau$ we compute analytical derivatives. For the derivatives of the lower row in Equation \ref{eq:system_model} with respect to the state $\vx$, we use numerical differentiation.

\subsection{Contact Model}
\begin{figure}[htbp]
\centering
\includegraphics[width=\columnwidth]{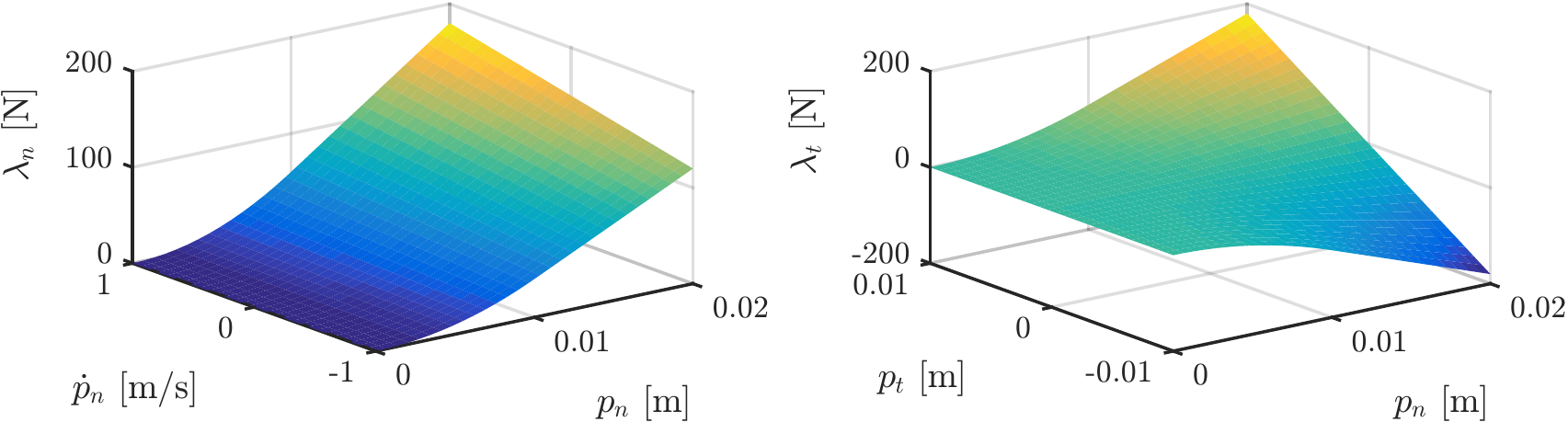}
\caption{Visualization of the contact model. The left plot shows the contact force in surface normal direction as a function of penetration and its derivative. The right plot shows the force in tangential direction as a function of penetration and tangential displacement. Both contact models are smooth at impact, i.e. at penetration 0.}
\label{fig:contact_models}
\end{figure}

Choosing or designing an appropriate contact model is critical for the performance of Trajectory Optimization. There are two main criteria defining the performance of a contact model: Physical accuracy and numeric stability. For Trajectory Optimization it is beneficial to use a smooth contact model which provides good gradients of the dynamics. Yet, this can lead to undesirable, unphysical defects like ground penetration. Therefore, there is a trade-off to be made. To mitigate this trade-off we are using a non-linear spring-damper contact model extending the model proposed in \cite{slq}. We consider two contact models: One collinear and one orthogonal to the surface normal. The normal contact model is defined as
\begin{equation}
    \lambda_N = 
        \begin{cases} 
            0                                & p_n \leq 0 \\
            (k_n + d_n \dot{p}_n)  \frac{p_n^2}{2\alpha_c} \vn_s       & 0 < p_n < \alpha_c \\
            (k_n+d_n \dot{p}_n) (p_n - \frac{\alpha_c}{2}) \vn_s          & p_n \geq \alpha_c
        \end{cases}
\end{equation}
while the tangential model is defined as 
\begin{equation}
    \lambda_t = 
        \begin{cases} 
            0                                & p_n \leq 0 \\
            (k_t p_t \vn_{d} + d_t \dot{p}_t \vn_{v})  \frac{p_n^2}{2\alpha_c}        & 0 < p_n < \alpha_c \\
            (k_t p_t \vn_{d} + d_n \dot{p}_n \vn_{v}) (p_n - \frac{\alpha_c}{2})           & p_n \geq \alpha_c
        \end{cases}
\end{equation}
Both models include a proportional and derivative terms which can be imagined as springs with stiffnesses $k_n$ and $k_t$ and dampers with damping ratios $d_n$ and $d_t$ respectively. For the normal direction the spring displacement $p_n$ is defined as the ground penetration orthogonal to the surface normal $\vn_s$. In tangential direction, the offset $p_t$ is computed between the current contact location and the location where the contact has been established. This effectively creates additional states in our system. However, we treat these states as hidden states, such that the size of the SLQ problem remains constant. In case of the tangential model, the force vector $\lambda_t$ is composed of the spring force in tangential displacement direction $\vn_{d}$ and a damping force in displacement velocity direction $\vn_{v}$. Both, the normal and tangential contact models are visualized in Figure \ref{fig:contact_models}.

\begin{table}[htbp]
\centering
\caption{Typical Values for Contact Model Parameters}
\label{tbl:contact_parameters}
\begin{tabular}{|c|c|c|c|c|}
\hline
$\alpha_c$ & $k_n$          & $d_n$          & $k_t$        & $d_t$         \\
\hline
0.01     & 8000 ... 90000 & 2000 ... 50000 & 0 ... 5000000 & 2000 ... 5000 \\
\hline
\end{tabular}
\end{table}

To avoid discontinuities during contact changes, the models are non-linear towards zero ground penetration, i.e. for $p_n < \alpha$, where $\alpha_c$ is a smoothing coefficient. Typical values for the contact model are given in Table \ref{tbl:contact_parameters}. The parameters are task dependent to trade-off physical correctness and convergence.

Friction and friction limits are considered via friction cones. This allows the Trajectory Optimization algorithm to reason about possible slippage and contact force saturation. In this work, we assume that static and dynamic friction coefficients are the same. Therefore, the friction cone can be expressed as a contact force saturation
\begin{equation}
    \vF_t = max(||\vF_t||, \mu F_n) \vn_s
\end{equation}
where $F_t$ is the contact force parallel to the surface, $F_n$ is the force normal to the surface, $\mu$ is the friction coefficient and $\vn_s$ is the surface normal.

\section{State Estimation and Tracking Control}
\begin{figure}[htbp]
\centering
\includegraphics[width=\columnwidth]{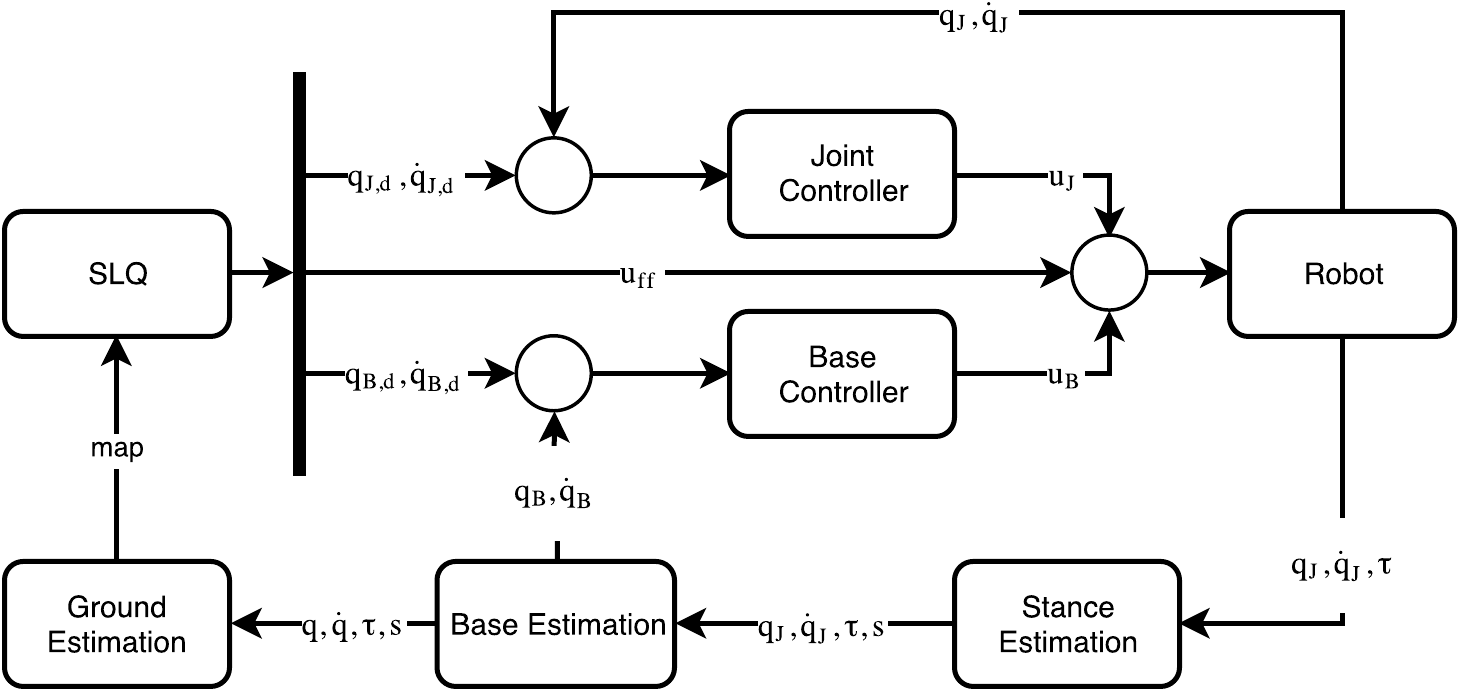}
\caption{Overview of the control and estimation pipeline. Base, stance and ground estimators provide information about the location and contact configuration of the robot. A joint controller and a base controller jointly control the robot's state.}
\label{fig:control_overview}
\end{figure}

Our TO approach is fully integrated into our estimation and control framework, illustrated in Figure \ref{fig:control_overview}. Since SLQ assumes full state-feedback, joint position and velocity measurements are fused with IMU data to obtain a base estimate\cite{bloesch2013state}. Furthermore, since SLQ reasons about contacts with the environment, a ground estimator is added which estimates elevation and inclination of the ground. SLQ then optimizes a trajectory which is then fed to a task-space base controller and the joint controller. These control loops are closed at a rate of 200 Hz. The corrective output of both controllers is then added to the feedforward control signal as optimized by SLQ and sent to the robot.

\subsection{State Estimation}
SLQ assumes full state-feedback, i.e. all states are assumed to be known. While we can directly measure joint positions and velocities with the encoders on HyQ, the base state cannot be directly measured. Therefore, we use \cite{bloesch2013state} to estimate the base's pose and twist.

The contact state is estimated by mapping joint torques to estimated ground reaction forces. If the normal component of the ground reaction force estimate is above a fixed threshold we assume the leg being in contact. In order to reason about the contact forces $\lambda$, we also have to know the elevation and contact surface normal at the contact points. In order for the algorithm to work online we cannot assume a perfectly leveled ground. Instead, we assume a ground plane which is fitted to the last contact point of each feet. This implementation allows for incorporating a ground map in the future.

\subsection{Tracking Controller}
SLQ does not only optimize feed-forward control action but also a feedback controller. While we have successfully applied theses feedback gains to robotic hardware \cite{mpc_quad}, we are not using the optimized feedback gains in this work for two reasons. First, the optimized feedback gains correspond to a time-varying, linear-qaudratic regulator (TVLQR). Therefore, the gains significantly depend on the linearization point. Especially due to the non-linearity of the contacts, the robot's state will drift from the linearization point during trajectory execution. Secondly, due to the usage of a single cost function, any regularization in the cost function will also effect the feedback gains which is undesirable. One solution is to recompute the gains online. However, we have found that the following tracking controller shows great performance at reduced complexity.

Assuming a known, fixed stance configuration with no slippage and at least three stance legs, the stance legs as well as the torso can be described by a 6 degrees of freedom state. This means, the base state, consisting of pose and twist, is coupled with joint positions and velocities of the stance legs. Therefore, controlling either set of states, the joint states of the stance legs or the base state, will subsequently also track the other set. A base controller allows us to directly track the base state and tune feedback gains on these states intuitively. Yet, for swing legs, we still require a joint controller. Hence, to get best of both worlds, we are using a combination of a task space base controller and a joint space controller. The joint space controller tracks the desired position and desired velocity of all joints independently. Additionally, the base controller regulates the base state with a PD controller. This task space control can be imagined as virtual springs and dampers attached to the robot's trunk on one side and the optimized body trajectory on the other \cite{pratt2001virtual}. 
\begin{equation}
	\vF_{cog} = \vP_x (\vx_{cog}^* - \vx_{cog}) + \vD_x (\dot{\vx}_{cog}^* - \dot{\vx}_{cog})
\label{eq:virtual_model}
\end{equation}
The desired body wrench $\mathbf{F}_{cog}$ is applied to the robot by converting it to forces in the contact points $\boldsymbol{\lambda}_{c}$ and then mapping these to the joint torques through $\boldsymbol{\tau}_{fb} = \mathbf{J}_{c}^T \boldsymbol{\lambda}_{c}$. These torques are then added to the feedforward control action obtained from SLQ. This feedforward control action already includes torques that counteract gravity and thus, no gravity compensation term is added in Equation \ref{eq:virtual_model}.

\section{Experiments}
To evaluate the approach, we first show, how a galloping and a trotting gait with optimized contact switching sequences and timings can be obtained in simulation. Furthermore, we demonstrate a manipulation motion that includes contact switches and more complex base motions. To verify the applicability of our approach, we demonstrate hardware experiments for the last task. Finally, we show a highly under-actuated task where HyQ walks on its hind legs like a humanoid. For each task, the cost function is shown as a color code below the heading. The colors indicate the individual, relative weightings of the diagonal entries of each weighting matrix, ranging from lowest (green) to highest (red) on a logarithmic scale. No color means that the according value is zero and all off-diagonal elements are zero as well. The input cost matrix $\vR$ is set to identity in all experiments. All tasks are initialized with a simple PD feedback controller on all joints and no feedforward control. This leads to HyQ maintaining its initial state over the entire length of each trajectory.

\subsection{Simulation}
\begin{figure}[htbp]
\centering
\includegraphics[width=\columnwidth]{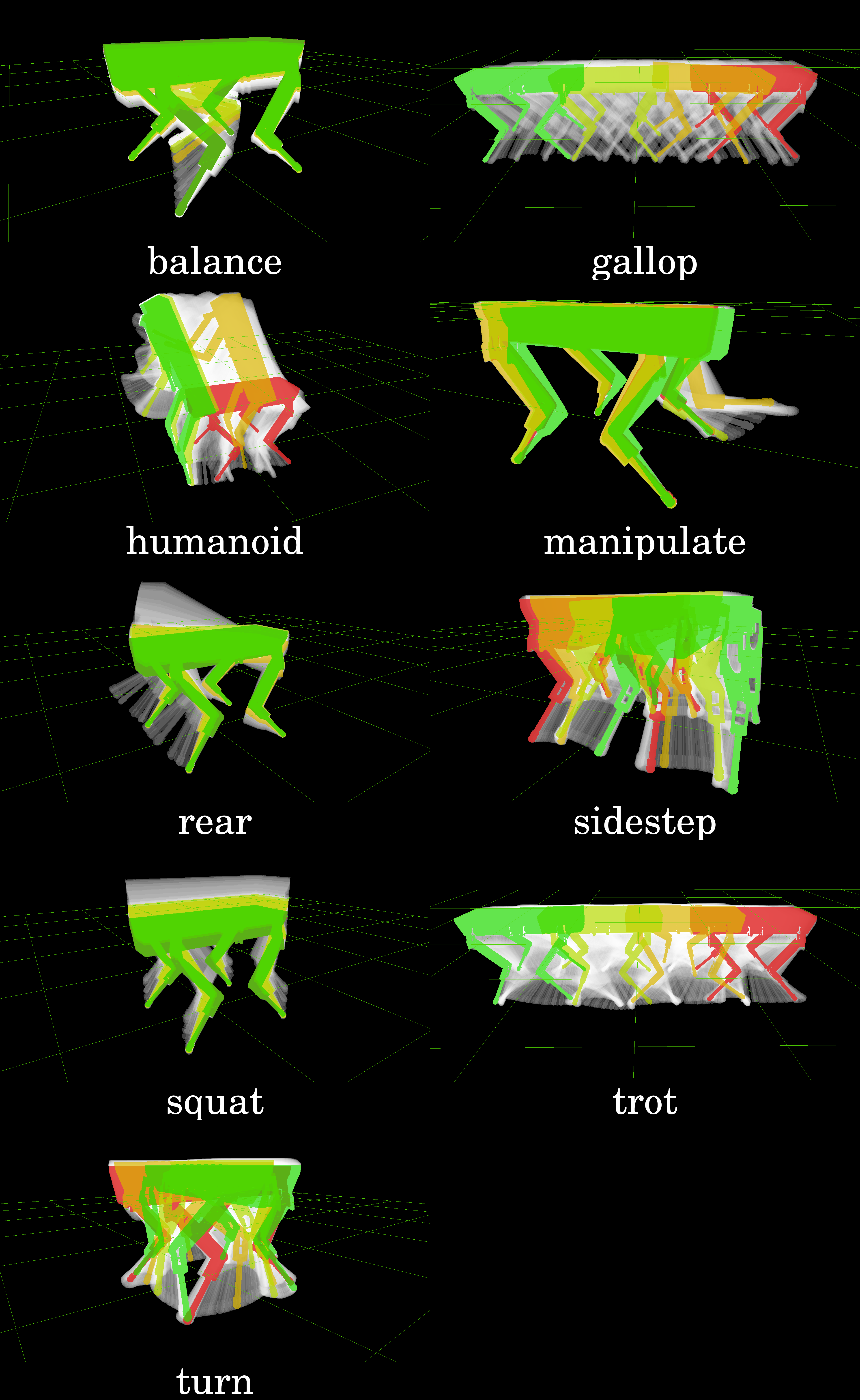}
\caption{Time series of the optimized trajectories for each task. Poses are shown as a color gradient over time ranging from red (initial pose) to green (final pose). Intermediate poses are indicated in transparent white. All displayed motions contain contact switches during dynamic maneuvers. These contact switches result from optimization and are not pre-specified.}
\label{fig:strips}
\end{figure}
\subsubsection{Galloping}

\begin{center}
    \includegraphics[width=\columnwidth]{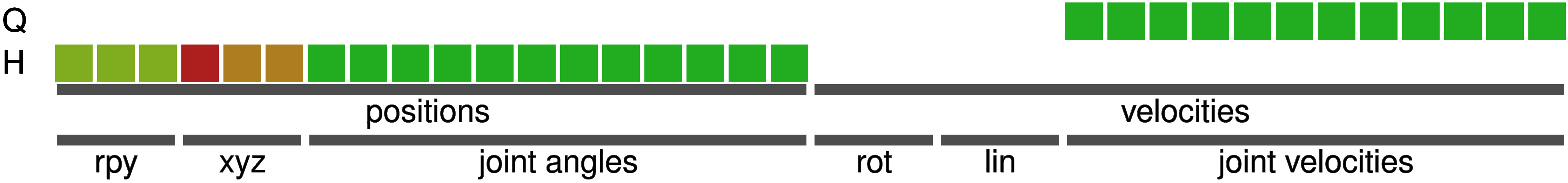}
\end{center}

The first gait we demonstrate is galloping. The galloping gait is a direct outcome of the optimization with a very simple cost function. The cost function consists of final costs on the base position and leg positions. Additionally, we add some regularization on the base and leg motion to prevent excessive motions of the body or the limbs. However, the cost function does not include any terms related to contact sequences or timings and no priors on a galloping gait. Also, we are not using any intermediate cost terms here. HyQ starts in its default stance and is supposed to reach its final position 2 m in front. The chosen time horizon is 3 seconds. As for all tasks, the initial controller is a simple joint PD controller which results in HyQ staying in place and maintaining its intial configuration.

\begin{figure}[htbp]
\centering
\includegraphics[width=\columnwidth]{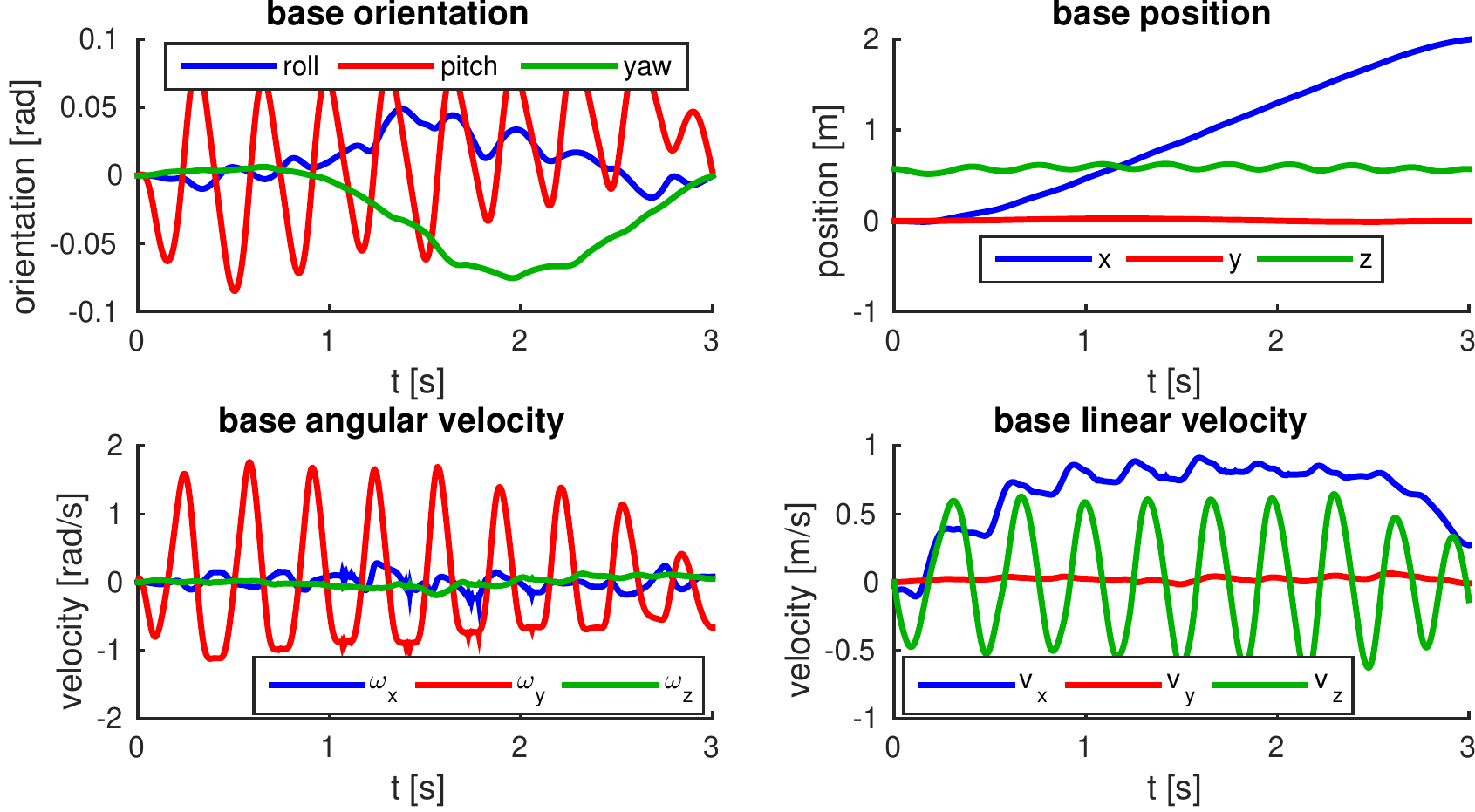}
\caption{Plots of base pose and base twist during galloping as optimized by SLQ. The robot takes in total 9 galloping steps. As expected, there is significant pitch motion during galloping. The desired final position at $x = 2.0$ m is reached with good accuracy. }
\label{fig:gallop_base}
\end{figure}
\begin{figure}[htbp]
\centering
\includegraphics[width=\columnwidth]{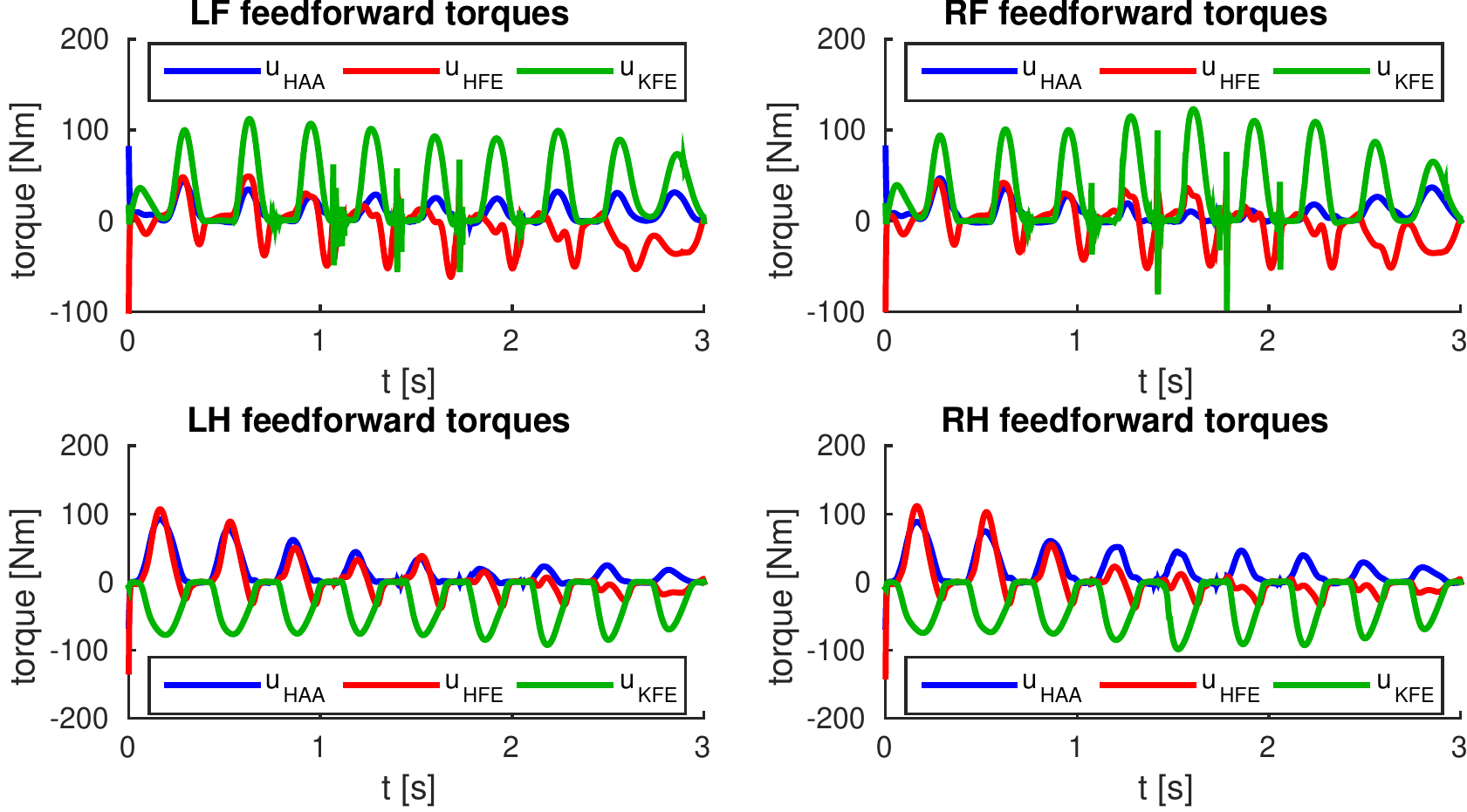}
\caption{Torque profiles of the different joints during a galloping motion. As the lower plots show, the hind legs and especially the HFE joints, contribute greatly to the acceleration. The front leg torques seem to contribute fairly evenly to the galloping motion throughout the trajectory.}
\label{fig:gallop_torques}
\end{figure}

As the results in Figure \ref{fig:gallop_base} illustrates, we obtain a gallop motion with 9 steps which includes acceleration and deceleration. Finally, the robot reaches the desired position at $x=2.0$ m. As expected, we see significant pitch motion of the upper body. From Figure \ref{fig:gallop_torques} we can tell that the hind legs, especially the HFE joints, are used for acceleration. While our findings are not directly comparable to biology, we can see similar effects: In fast running animals, like leopards and horses, the rear legs also bear a larger load during running than the front legs, which explains the different sizing of legs and muscles between rear and front. Snapshots of the gallop are shown in Figure \ref{fig:strips} and the full motion is shown in the video\footnote{\label{video}\url{https://youtu.be/sILuqJBsyKs}}.

\subsubsection{Trotting}
\begin{center}
    \includegraphics[width=\columnwidth]{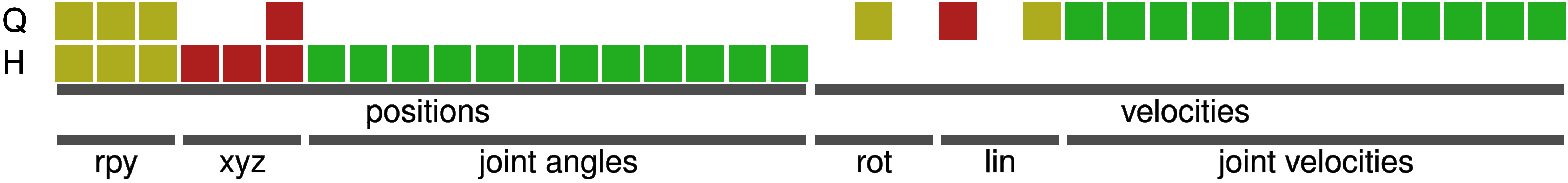}
\end{center}

One hypothesis for the previous task resulting in a galloping behavior is the short time horizon and no penalty on the orientation. Now if we increase the time horizon to 8 seconds and penalyze the base motion, i.e. we give HyQ more time and encourage smoother base motions, we see that the optimization starts to prefer a trotting motion instead of the galloping. Interestingly enough, the weightings allow to blend between galloping and trotting. As a result, HyQ first gallops and then smoothly transitions to a trotting gait. Again, neither the contact sequence, nor the gait or the transition is given. The parameters influencing the gait are the diagonal elements of $\vQ$ and $\vR$ as in Equation \ref{eq:cost_function_experiments}.

By increasing the base motion penalty, the trajectory results in a pure trot. The trot consists of 4 steps per diagonal leg pair with almost constant stride length. By setting the desired position to the side instead of the front, the resulting trajectory is a sidestepping motion. If only a desired yaw angle is set, the robot turns on the spot. Both trajectories are trotting variants where the diagonal leg pairs are moved together. Both motions are illustrated in Figure \ref{fig:strips} and included in the video\textsuperscript{\ref{video}}.

\subsubsection{Squat Jump}

\begin{center}
    \includegraphics[width=\columnwidth]{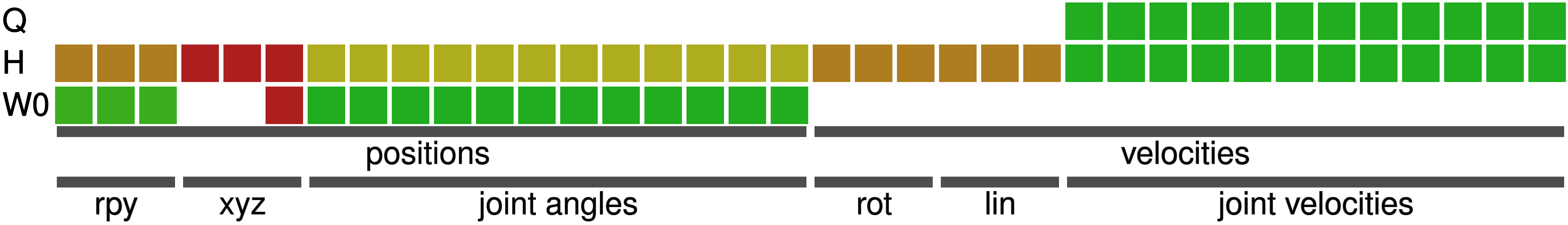}
\end{center}

\begin{figure}[htbp]
\centering
\includegraphics[width=\columnwidth]{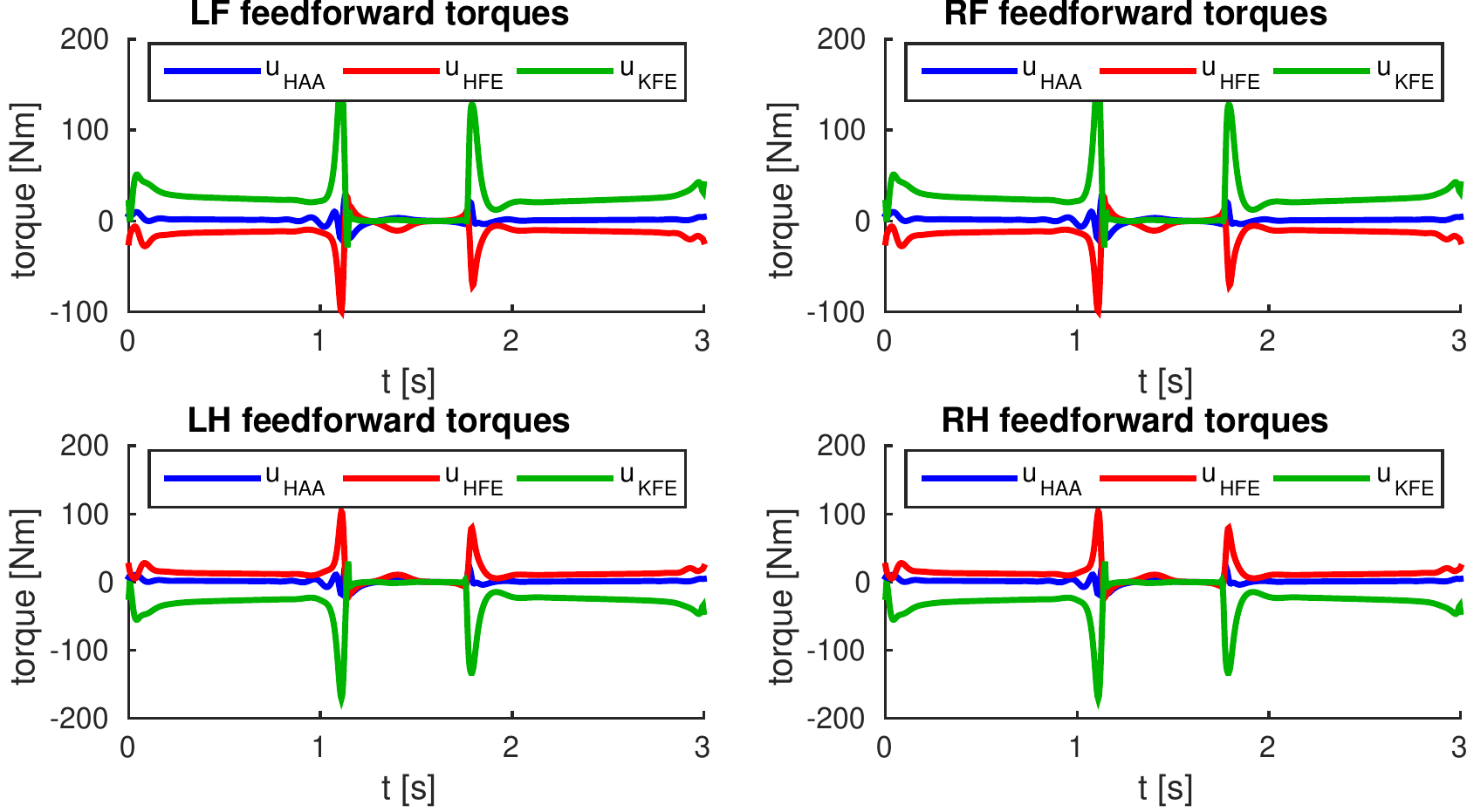}
\caption{Plots of the torque profiles during a squat jump. During stance phases the profiles are fairly flat. There are two distinct spikes which are the torques produced during take-off and landing. By including the torques in the optimization criterion of the squat jump task, the distribution is even between all legs.}
\label{fig:squat_torques}
\end{figure}

Next, we test if our Trajectory Optimization approach can leverage and reason about the dynamics of our system. Thus, we define a task with intermediate waypoint states that are not statically reachable. First, we add an intermediate waypoint cost term for the base position and orientation. For this waypoint, we choose a strong weighting for the base height and set the desired value of the base height to $z=0.8 m$. Furthermore, the waypoint costs contain low weighting on the base orientation to keep the base level. By adding a cost on the deviation from default joint position, we ensure HyQ cannot just try to straighten its legs to try to reach the base pose, but that it has to jump. After running our optimization, we obtain a near symmetric squat jump. The overall optimization spans the entire motion, e.g. preparation for lift-off from default pose, the lift-off itself, going to default pose in the air, landing and returning to the default pose. The apex waypoint is localized in time but contact switches and timings are optimized. Also, we are not directly specifying a squat jump. Therefore, simultaneous lift-off and landing of the legs is an optimization outcome rather than pre-specified.
\begin{figure}[htbp]
\centering
\includegraphics[width=\columnwidth]{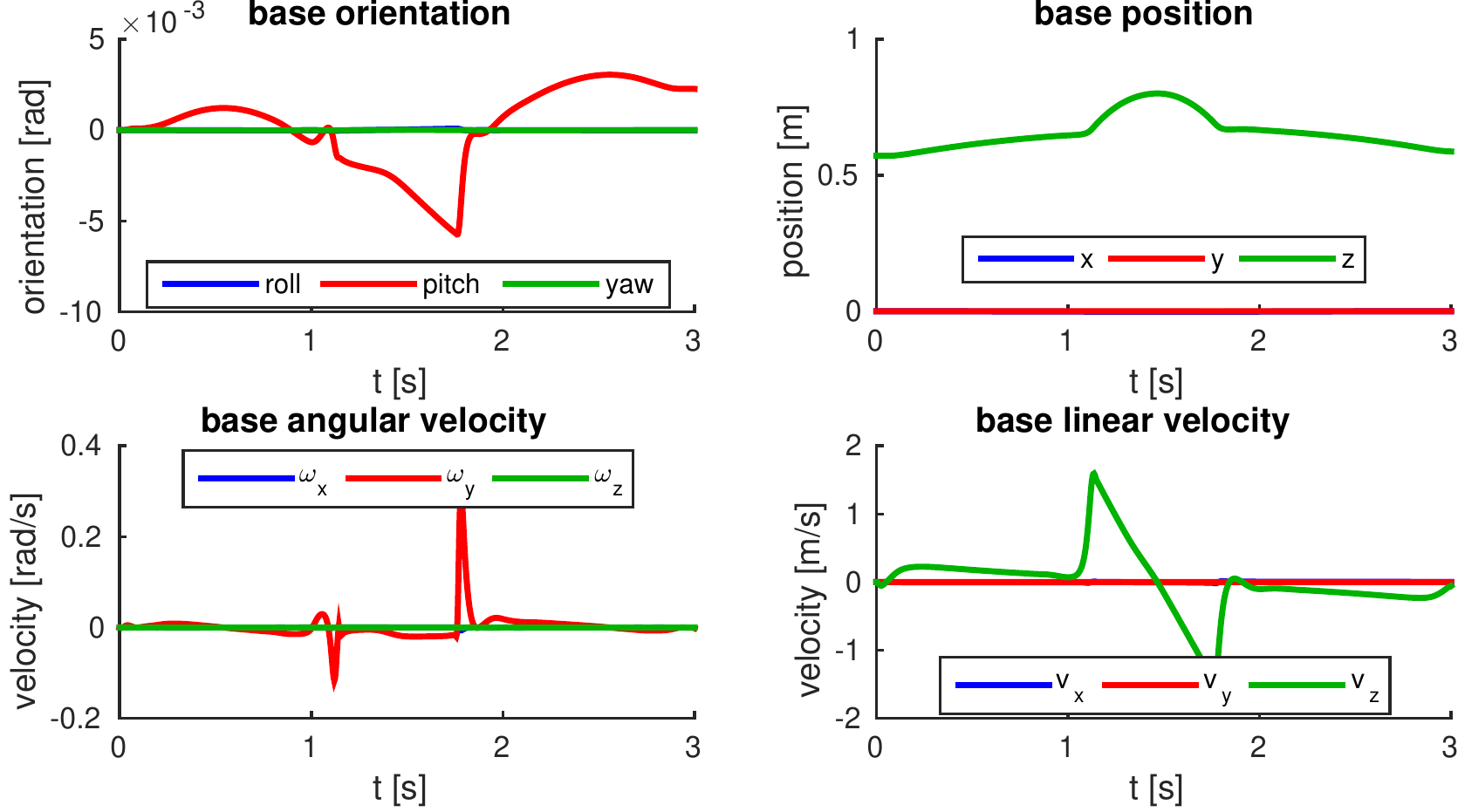}
\caption{Plots of base pose and base twist during a squat jump as optimized by SLQ. The graphs show that the robot keeps its base stable while jumping. Also unnecessary base motion in x- and y-direction are minimized. Furthermore, we can see that the desired apex height of 0.8 m is reached (measured value 0.8004).}
\label{fig:squat_base}
\end{figure}
Figure \ref{fig:squat_base} shows the optimized squat jump motion. We can see that unnecessary motions such as rotations or position changes in x- or y-direction are avoided. Also, the desired apex height of 0.8 m, despite being a soft constraint, is reached with sub-millimeter accuracy (measured 0.8004 m). The small pitch velocity results from an off-center mass and the optimization goal of low torques which promotes equal distribution of torques between legs as seen in Figure \ref{fig:squat_torques}. The same plot also shows that the largest torques appear in the KFE and HFE joints. Furthermore, the lower leg of HyQ weighs less than 1 kg compared to a total mass of about 80 kg. Therefore, the mass that the lower leg drives changes rapidly during contact changes occurring at around $t=0.55 s$ and $t=0.9 s$. Hence, the torques also change quickly. While the knee joint is still moved in the air to take the default angle and prepare for landing, the required torque is minimal. Again, both motions are shown as a snapshot series in Figure \ref{fig:strips} as well as in the video\textsuperscript{\ref{video}}.
%

\subsubsection{Rearing}

\begin{center}
    \includegraphics[width=\columnwidth]{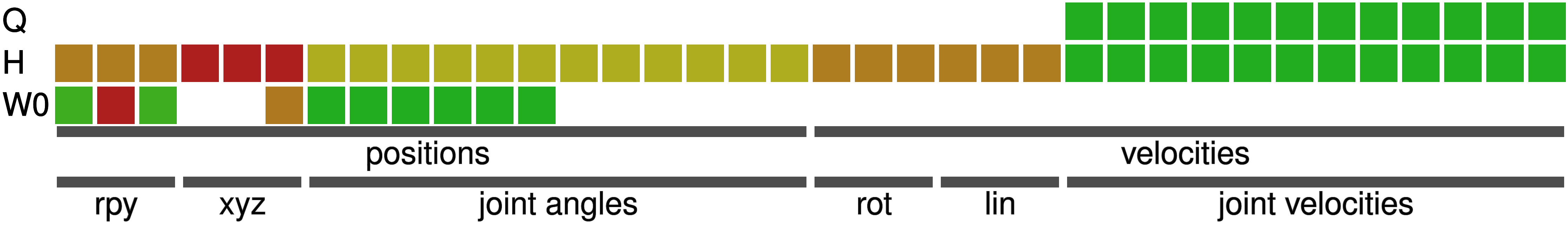}
\end{center}

For the next task, we are using a similar cost function as the squat jump. Instead of penalizing the deviation from a neutral base pose, we penalize deviations from a 30$^\circ$ pitched base orientation and we lower the desired apex height to 0.7 m. Yet again, we do not specify contact/stance configurations. Therefore, the optimization algorithm is free to optimize the trajectory. The final trajectory represents a rearing motion, where HyQ lifts off with the front legs, reaches the apex position and finally returns to full contact as well as its default pose. A screenshot of the apex position is shown in Figure \ref{fig:strips}. The full motion is shown in the accompanying video\textsuperscript{\ref{video}}.

\subsubsection{Diagonal Balance}
\begin{center}
    \includegraphics[width=\columnwidth]{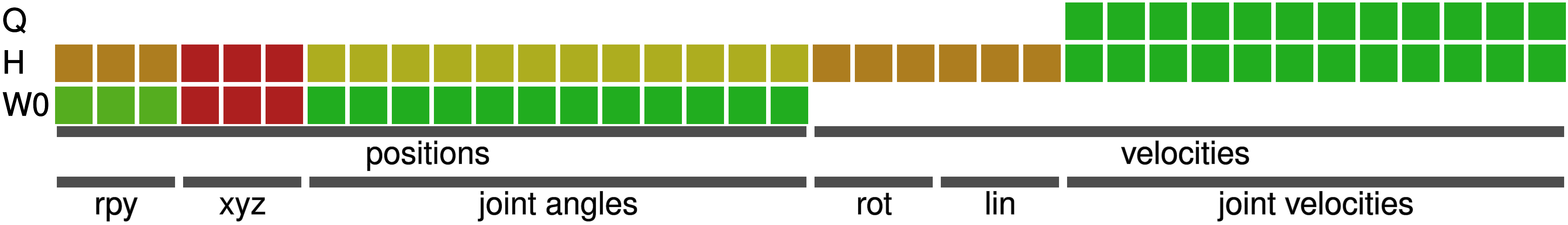}
\end{center}

In order to demonstrate that the SLQ can also find statically unstable trajectories, we are demonstrating a diagonal balance task. Here, we use a waypoint term in our cost function again. This term penalizes the orientation and height of the base. Furthermore, it encourages the robot to pull up its legs by bending HFE and KFE of the left front and right hind leg. The final trajectory shows the expected balancing behavior. Again a screenshot at apex is shown in Figure \ref{fig:strips} and the video\textsuperscript{\ref{video}} shows the full motion. Interestingly, while we are using a single intermediate term with a single time point and absolutely symmetric costs, the lift-off and touch-down of the swing legs is not synchronous but the front left leg lifts-off later and touches down earlier. This asymmetry most likely stems from the asymmetric location of the CoG and assymmetric inertia of the robot's main body.

\subsubsection{Humanoid Walk}

\begin{center}
    \includegraphics[width=\columnwidth]{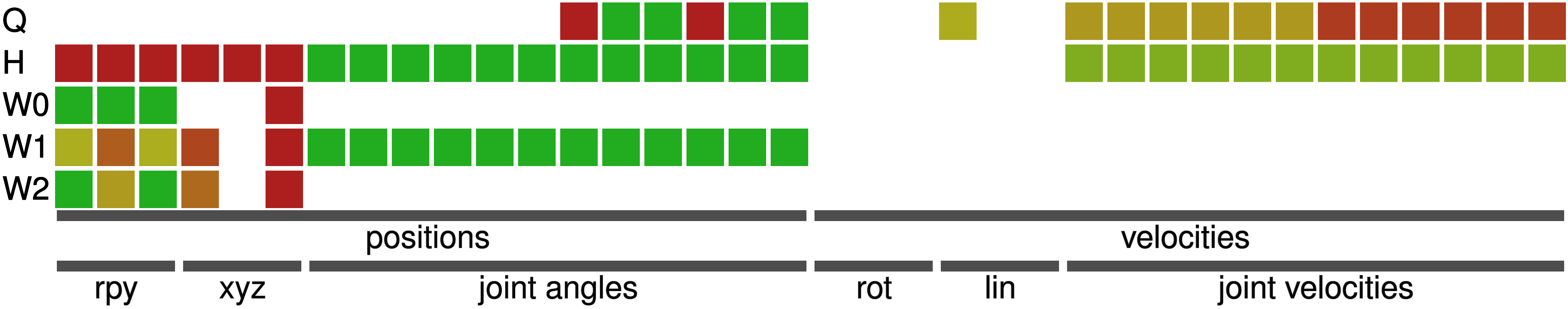}
\end{center}

As a final test, we evaluate if our algorithm is capable of optimizing a behavior that allows HyQ to stand on its hind legs, similar to a humanoid robot, and then go to a target point in front. While standing on its hind legs exceeds the capabilities of the hardware, such a motion is physically possible. Compared to a humanoid with ankles, HyQ has point feet. This increases the level of difficulty since the absence of ankles makes HyQ highly under-actuated. This means that during a standing or walking task no torques can be applied to the ground but balance can only be maintained by moving the links.

In this task, we again initialize HyQ in its default stance, requiring it to stand up in place first and then reach a goal point where it should stabilize for a second. The first waypoint has a very wide spread in time and penalizes deviations in base orientation and height. This cost ensures that HyQ stays upright during the entire task after getting up. We do not add this cost to the general intermediate cost, as it would encourage HyQ to stand up as fast as possible, rather than getting up in a controlled, efficient manner. The stand up motion is encouraged by the second waypoint cost penalizing base orientation, height and changes in forward position. We add a third waypoint one second before the end of the time horizon, defining the target pose and orientation. While the last waypoint and the final cost specify the same base pose, we separate them to demonstrate that HyQ can stay upright and stabilize in place for longer times without showing signs of falling.

The resultant motion shows interesting, non-trivial behavior. Before getting up HyQ pulls its hind left leg in, moving the contact point more closely below the center of gravity. Also, it uses the front left leg (``left arm'') to get up, resulting in a very natural, coordinated, asymmetric motion. After getting into a two-leg standing phase, a forward motion is initiated by a short symmetric hopping but quickly changes to a coordinated stepping/walking pattern. While we penalize joint velocity, the front limbs are moved to support balance during the motion until they are retracted to a target pose at the end of the trajectory. The entire motion underlines once more the capabilities of the approach. Classic controllers and motion planners might not find such a complex motion. Additionally, by allowing optimization over contact points and sequences, we allow the Trajectory Optimization to find the best solution for the task. While a rearing motion, as demonstrated above, is feasible and can be found using our problem formulation, the optimizer prefers a less dynamic, coordinated motion here.

\subsection{Hardware}
\begin{center}
    \includegraphics[width=\columnwidth]{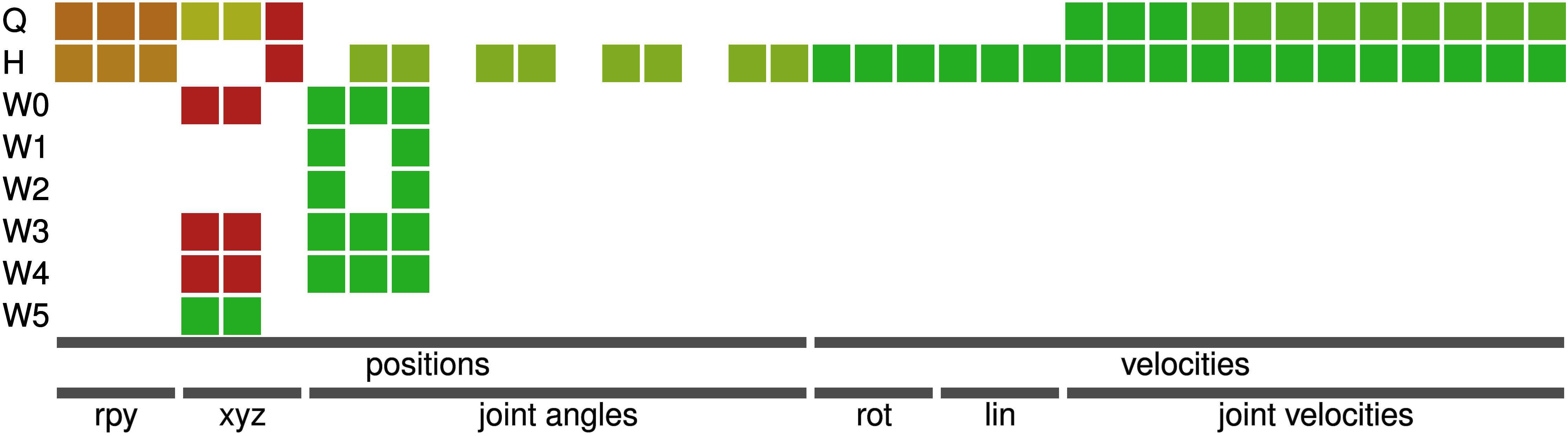}
\end{center}

To verify that the optimized trajectories can be applied to hardware, we run an example task on hardware. This task is a rough manipulation task where HyQ pushes over a pallet with its front left leg. The trajectory requires a sequence of motions such as shifting its CoG in the support polygon of the three stance legs, lifting off of the front left leg, executing the push motion, putting the foot back down and shifting the CoG back. While in classic approaches these motions would possibly all handcoded, they result from a single cost function with only one intermediate cost function term for the front left leg in our Trajectory Optimization approach. While this trajectory works perfectly fine in simulation, we have to slightly alter it for the hardware. Since our TO approach is deterministic and we penalize control input, the algorithm tries to minimize the shift of the CoG, leading to a very risky trajectory. Thus, small model inaccuracies or disturbances make the robot loose its balance. This is a general problem of all deterministic TO approaches and can only be fundamentally solved by using risk aware or stochastic approaches. While there is a risk-aware SLQ variant \cite{ileg2015}, here we apply a work-around by simply encouraging a larger CoG shift in our cost function. For visual purposes only, we add additional intermediate waypoints for the front left leg only, leading to a slower, more appealing push motion. While we could easily add the push contact to our optimization, we leave it unmodelled on purpose such that it becomes a disturbance to our controller, underlining its robustness. The pallet is a 1200 by 800 mm Euro pallet that weighs approximately 24 kg and thus requires a force of about 70 N at the point of contact of the leg to be pushed over.

\begin{figure}[htbp]
\centering
\includegraphics[width=\columnwidth]{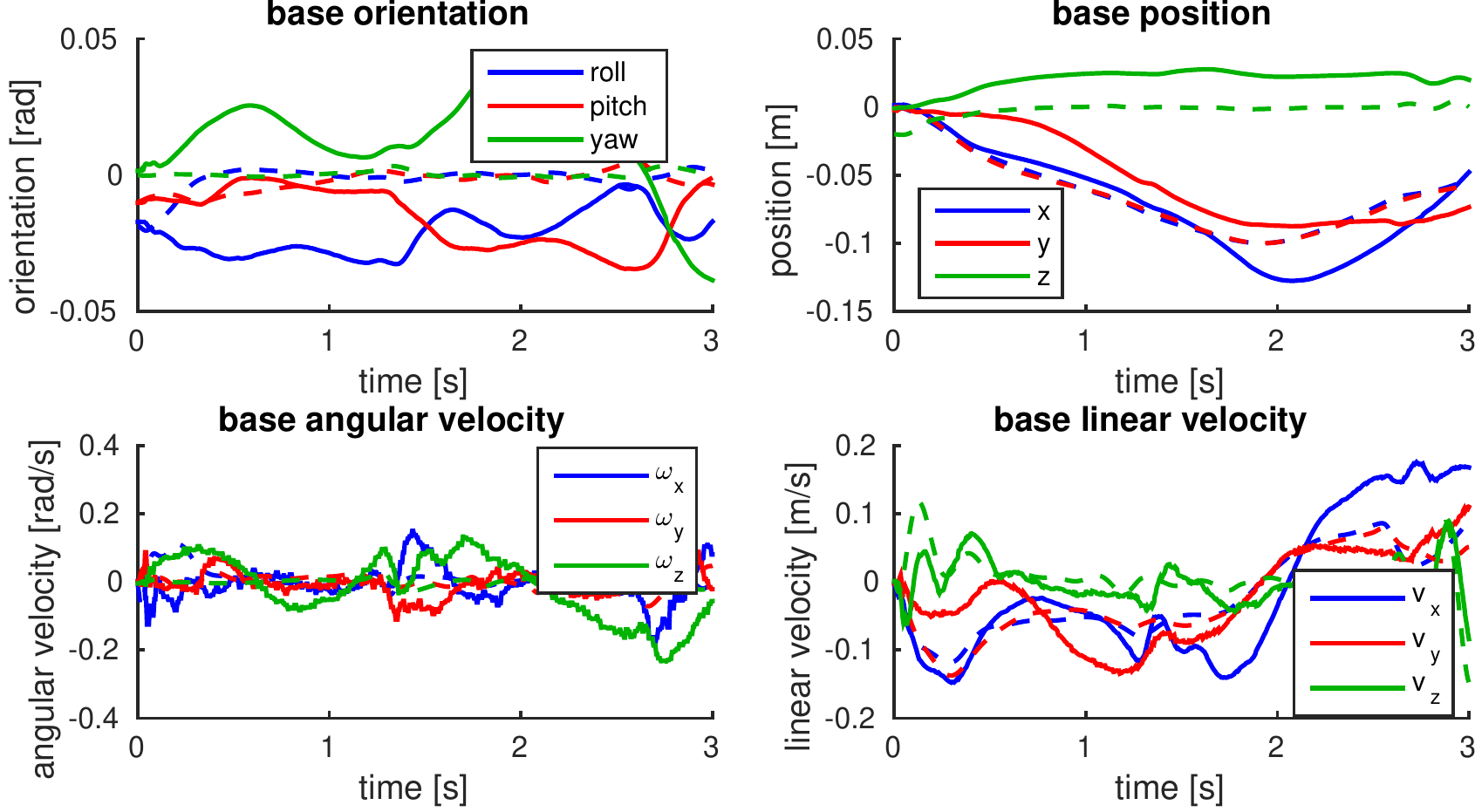}
\caption{Base pose and twist trajectories (solid) and their respective references (dashed) for the hardware test where HyQ is executing a rough manipulating task. The plots show that the planned and executed base trajectories are very similar. While the task space controller is tracking the base trajectory, it also gets indirectly tracked through the joint space controller.}
\label{fig:hw_base_tracking}
\end{figure}
\begin{figure}[htbp]
\centering
\includegraphics[width=\columnwidth]{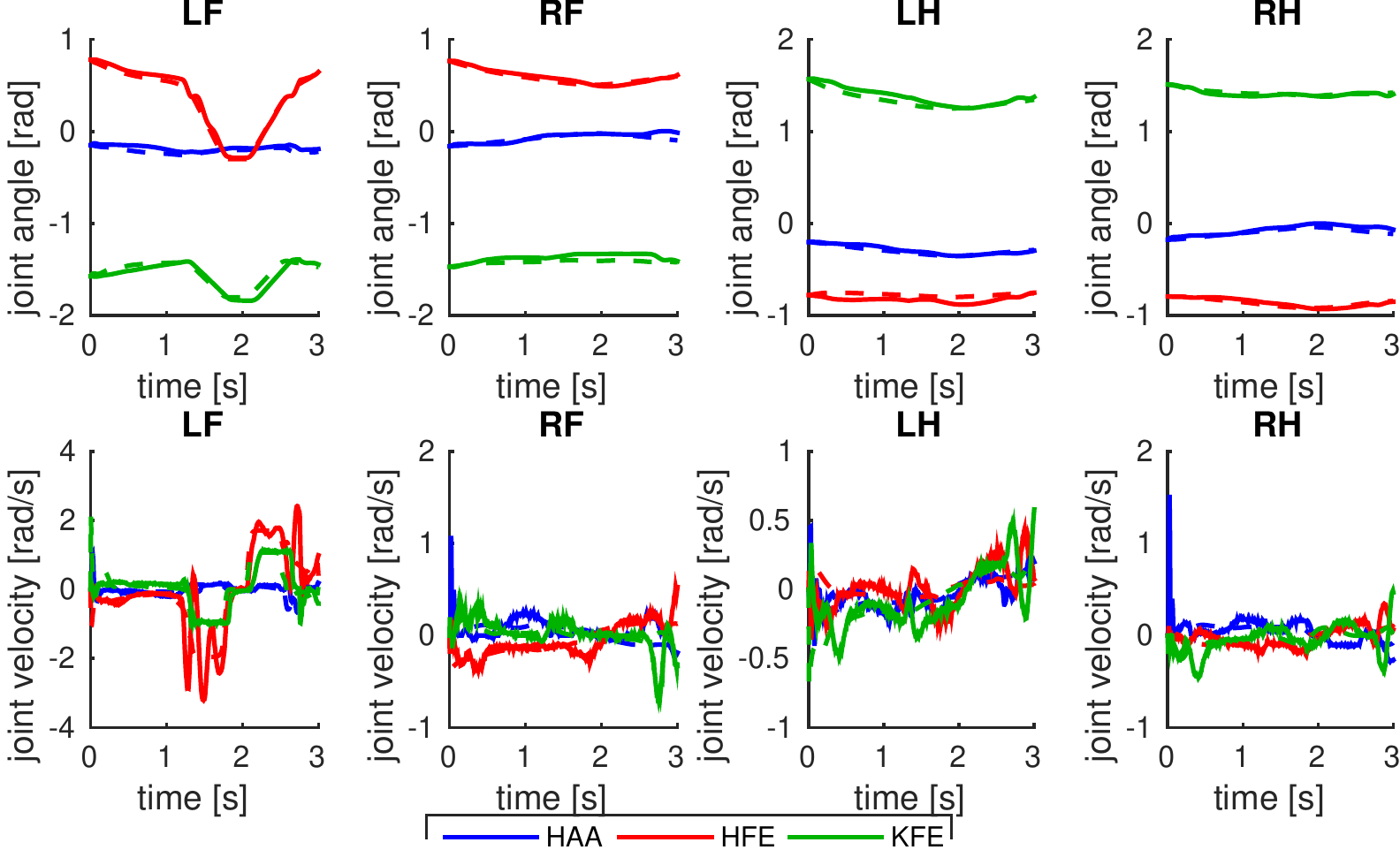}
\caption{Planned (dashed) and executed (solid) joint trajectories during the hardware test where HyQ is executing a rough manipulation task. The joint tracking controller follows the reference very well and thus also contributes to tracking the base trajectory.}
\label{fig:hw_joint_tracking}
\end{figure}
\begin{figure}[htbp]
\centering
\includegraphics[width=\columnwidth]{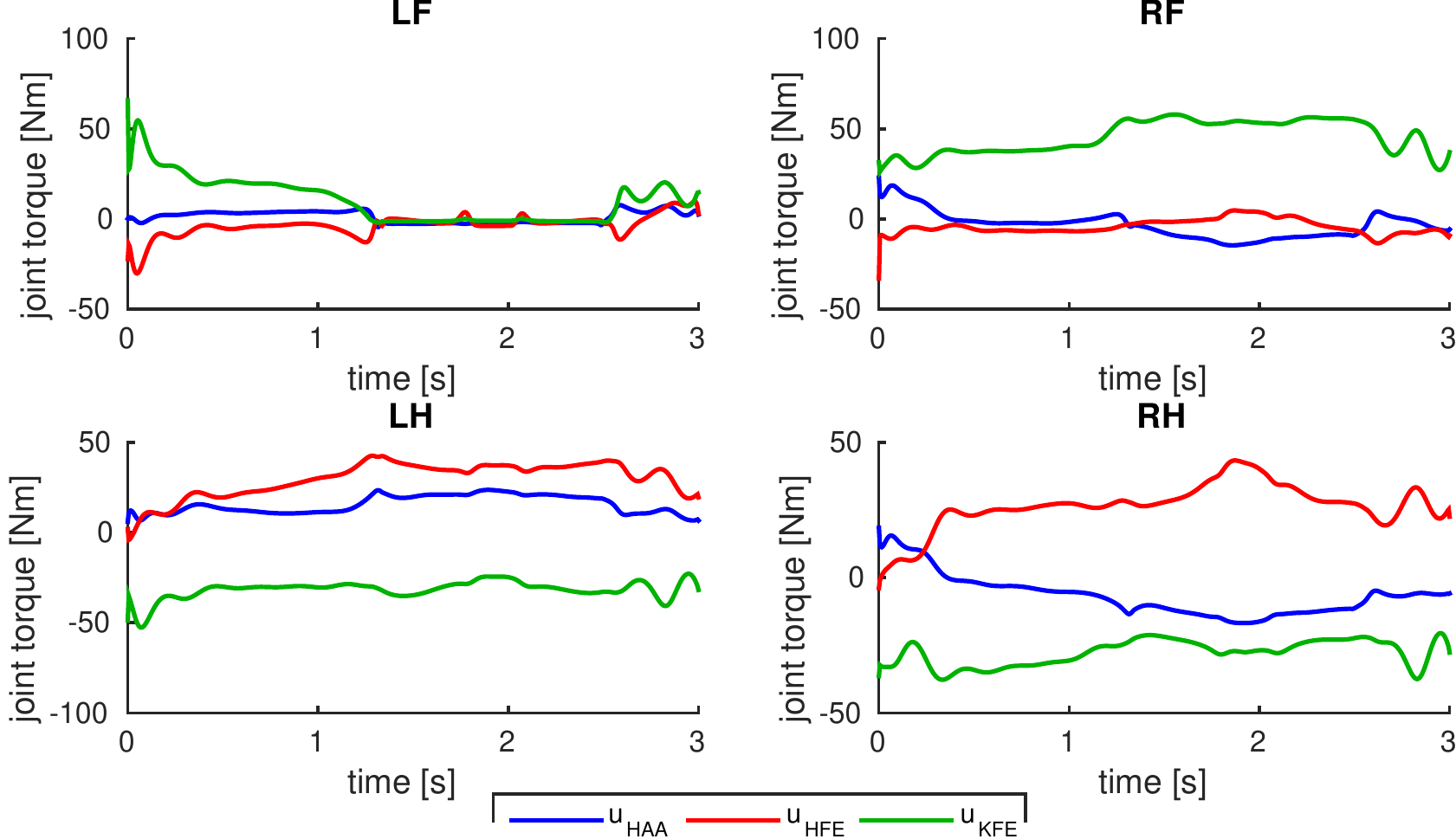}
\caption{Joint torques during the hardware test where HyQ is executing a rough manipulation task. These joint torques are the sum of the feedforward torques of the TO algorithm and the task space controller output. These torques are given to the joint space controller which adds additional tracking torques.}
\label{fig:hw_torques}
\end{figure}
To demonstrate that the algorithm is being run online we approach the pallet with a separate walking controller. Thus, the robot is faced with different initial conditions every time. Figure \ref{fig:strips} shows a sequence of images of the optimized motion while Figure \ref{fig:hw_sequence} shows a sequence of images taken during execution. The video\textsuperscript{\ref{video}} shows the entire motion as well as examples from our physics simulator, where initial conditions differ even more significantly. As can be seen in all examples, the motion is dynamic and lift-off and touch-down events of the front left leg partially overlap in time with the CoG shift. This underlines that static stability is not required in this task but the algorithm finds a dynamically stable trajectory.

The base pose/twist tracking for this task during hardware experiments is shown in Figure \ref{fig:hw_base_tracking} and the joint position/velocity tracking in Figure \ref{fig:hw_joint_tracking}. We can see that the joint tracking is better than base tracking due to a more aggressive joint controller. However, the base pose and twist do not significantly deviate from their planned trajectories either. Figure \ref{fig:hw_torques} shows the combination of feedforward torques and task space control input. In the front left leg, we can nicely see the lift off and touch down events in between which joint torques are almost zero. This happens since this leg does not further contribute to sustaining the weight of the body. Subsequently, the other legs have to bear a higher load. Especially in the neighboring legs, i.e. the right front and left hind leg, we see an increase in torques. Yet, this increase is not very significant since the robot has already shifted its CoG towards the support polygon of the stance legs.

\subsection{Runtime and Convergence}
When running Trajectory Optimization online, runtime and convergence become a major concern since these measures define how long the robot ``thinks'' before executing a task. Especially in a dynamic environment or in presence of drifting estimates, we want to keep the optimization procedures short to be able to react to a given situation quickly. Even better than pre-optimizing a trajectory before execution, we reoptimize and adjust them during execution, forming a model-predictive control scheme. In this section, we will look at both the number of iterations for each task as well as the runtime of each iteration. This gives us an indicator of the complexity of a task and tells us how far we are from running our approach in an MPC fashion.

\begin{figure}[!htbp]
\centering
\includegraphics[width=\columnwidth]{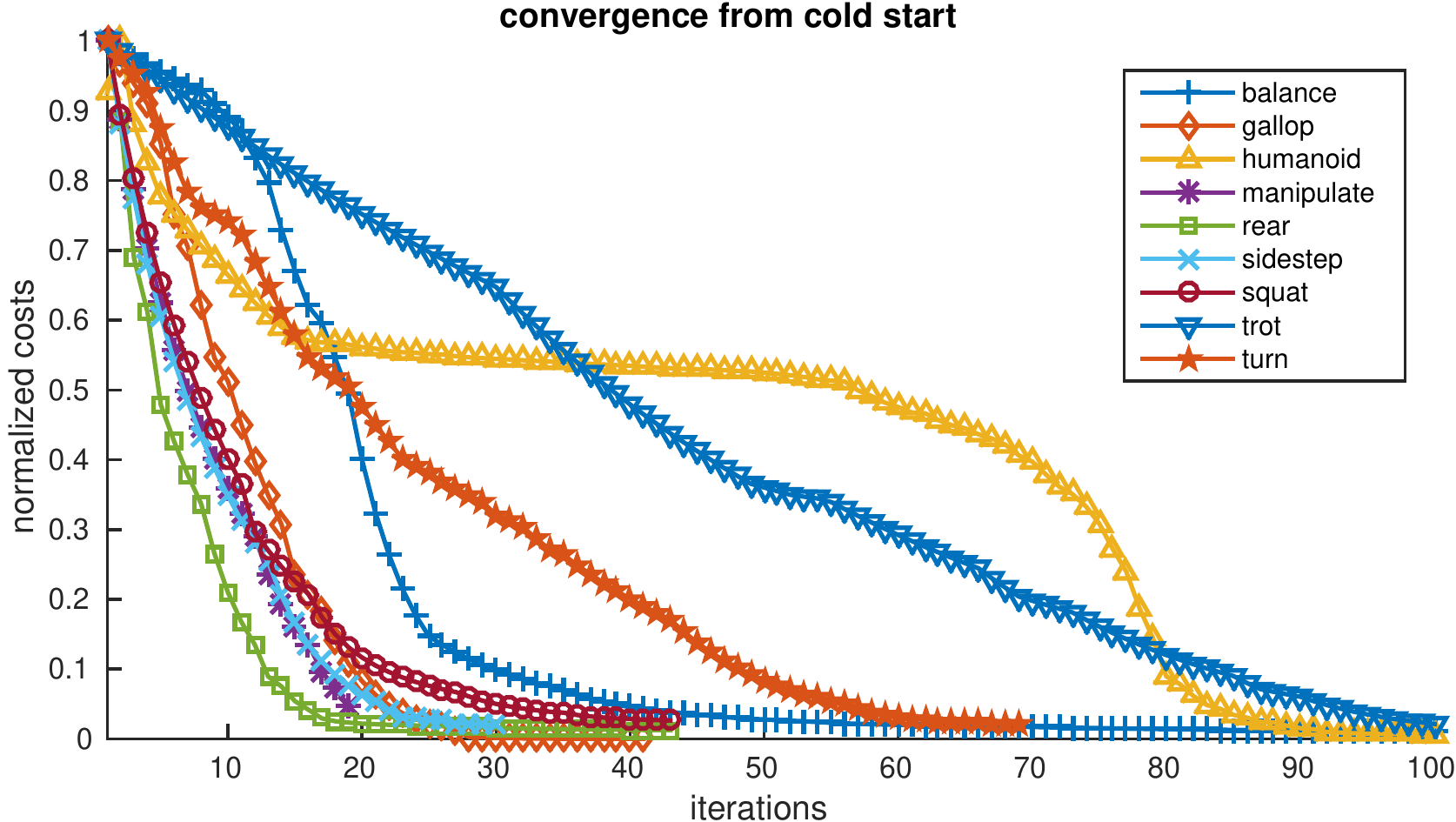}
\caption{Convergence rates for different tasks. The convergence rate seems to be influenced by the length and complexity of the task. The fastest convergence can be observed in the manipulation tasks since the complexity and duration is lower than in other tasks. The trotting behavior converges the slowest.}
\label{fig:convergence}
\end{figure}

\begin{figure}[!htbp]
\centering
\includegraphics[width=\columnwidth]{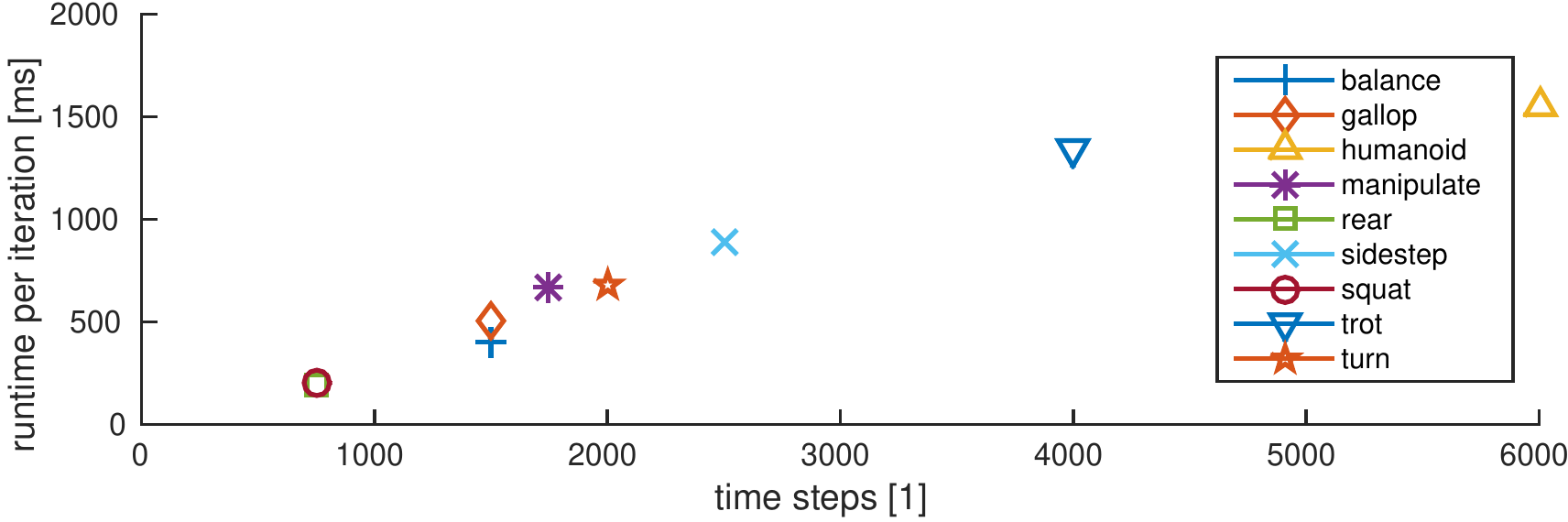}
\caption{Runtime per iteration for different tasks. From a theoretical point of view, the relation between number of time steps and the runtime per iteration should be linear. This plot nicely supports this hypothesis. As a result, SLQ scales well also for problems with large time horizons.}
\label{fig:runtime}
\end{figure}

As a first test, we take a look at convergence rates across tasks. To obtain comparable results between different tasks, we initialize all tasks with a standing controller in form of a pure joint position controller and normalize the costs with the initial cost. The results of this test are shown in Figure \ref{fig:convergence}. The curves suggest that there is a relation between the time horizon and complexity of a task and the corresponding convergence rate. The trotting behavior is one of the most complex behaviors and also has the longest time horizon. The rearing task is relatively simple and well guided by the waypoint costs which is a possible explanation of the fast convergence rate. Most tasks converge within 10 and 40 iterations, which given the runtime per iteration, shown in Figure \ref{fig:runtime}, usually means an overall optimization time of less than 1 minute. 

When it comes to runtime, SLQ has a major advantage over many other Trajectory Optimization approaches: It scales linear with discretization steps and thus with time horizon. Each SLQ step is fixed runtime which means the fewer steps we use, the faster the optimizer runs. Therefore, one would ideally increase step size but also shorten the trajectory length. Yet, there is a limit to both. Figure \ref{fig:runtime} underlines the linear relation between the number of trajectory points (or steps) and the runtime per iteration. The timings are measured on a standard quadcore laptop and averaged over multiple iterations. To achieve these runtimes we use a custom multi-threaded solver. Additionally, we use highly optimized, auto-generated code for the computation of forward dynamics, used for rollouts/line-search and linearization. This code is generated using RobCoGen \cite{robcogen}, a code-generator framework for rigid body kinematics and dynamics. 

\section{Discussion, Conclusion and Outlook}
In this paper, we have presented a fully dynamic, whole-body Trajectory Optimization framework that is able to create motions which involve multiple contact changes. The approach does not require any priors or initial guesses on contact points, sequences or timings. We demonstrate the capabilities on various tasks including squat jumps, rearing, balancing and rough manipulation. Furthermore, our approach is able to discover gaits such as galloping, trotting and two legged walking. First hardware results show that the optimized trajectories can be transferred to hardware. While some motions might not be directly applicable to hardware, they can serve as an initialization for other algorithms and they could potentially provide insight into optimal gait parameters such as stepping frequencies or stride lengths. As with most optimal control approaches, a certain amount of cost function tuning is required. However, we use only one cost function per task and tried to stay away from complex terms that tune for numerical properties. Instead, we tuned for functionality and behavior and provide insight into the weightings used. While tuning cost functions is a manual procedure, most of the obtained motion would require a complex combination of different planners and controllers if implemented with state-of-the-art approaches, e.g. \cite{victor_trot,coros2011locomotion}. In contrast to these methods, whole body Trajectory Optimization is versatile and the structure of the approach is not task dependent. Also, the optimized trajectories leverage the full dynamics of the system and do not rely on heuristics or simplified models. Using our formulation and solvers, the optimized trajectory is usually available in less than one minute, even for complex tasks and despite not pre-specifying contacts. This compares favorable to reported results in literature where solving time can be several minutes or even hours. 

However, there are also some short comings of the approach. The solution space of the approach is huge, e.g. we can apply our approach to a broad variety of tasks despite limiting ourselves to a single cost function structure. While this generality is a strength of the approach, it also requires the user to ``choose'' a solution by modifying cost function weights or adding additional term. While this is a more or less intuitive approach, one would wish to further reduce the number of open parameters by applying some form of meta algorithm.
Another drawback is that the optimized motions might not be very robust to model inaccuracies, disturbances or noise. Thus, alterations of the cost functions might be required to successfully execute these motions on hardware. This is a known issue of deterministic Trajectory Optimization approaches and can be tackled only by considering the stochasticity of the problem.
Lastly, SLQ cannot handle state constraints, which can be a limiting factor for complex Trajectory Optimization problems.

We believe that SLQ can be especially useful when applied as a shorter horizon MPC controller where its fast run-times are leveraged and complex state constraints are taken care of by a higher level planner/optimizer. While the run-time of SLQ seems far off from an MPC scenario, warm starting is an efficient measure to make the algorithm converge in only a few iterations. Also, run-time can be further reduced by limiting the time horizon. Thus, as a next step, we are planning to apply our approach in MPC fashion to HyQ, extending our work in \cite{mpc_quad}. Hopefully, SLQ-MPC will be able to act as a general stabilization and tracking controller as well as short horizon re-planner, allowing us to execute most tasks demonstrated in this work on hardware. One important step in this direction will be to include input constraints.

\section*{Acknowledgment}
This research has been funded through a Swiss National Science Foundation Professorship award to Jonas Buchli and the NCCR Robotics.



\bibliographystyle{IEEEtran}
\bibliography{bibliography/bib}
%


\end{document}

%% file: commands.tex
\newcommand{\myvec}[1]{\boldsymbol{#1}}

\newcommand{\vdx}{\delta\myvec{x}}
\newcommand{\vdu}{\delta\myvec{u}}

\newcommand{\vf}{\myvec{f}}
\newcommand{\vg}{\myvec{g}}

\newcommand{\vl}{\myvec{l}}

\newcommand{\vn}{\myvec{n}}

\newcommand{\vp}{\myvec{p}}
\newcommand{\vq}{\myvec{q}}
\newcommand{\vr}{\myvec{r}}

\newcommand{\vu}{\myvec{u}}

\newcommand{\vx}{\myvec{x}}

\newcommand{\vA}{\myvec{A}}
\newcommand{\vB}{\myvec{B}}
\newcommand{\vC}{\myvec{C}}
\newcommand{\vD}{\myvec{D}}

\newcommand{\vF}{\myvec{F}}
\newcommand{\vG}{\myvec{G}}
\newcommand{\vH}{\myvec{H}}

\newcommand{\vJ}{\myvec{J}}
\newcommand{\vK}{\myvec{K}}

\newcommand{\vM}{\myvec{M}}

\newcommand{\vP}{\myvec{P}}
\newcommand{\vQ}{\myvec{Q}}
\newcommand{\vR}{\myvec{R}}
\newcommand{\vS}{\myvec{S}}

\newcommand{\vW}{\myvec{W}}














%% file: main.bbl
\begin{thebibliography}{10}
\providecommand{\url}[1]{#1}
\csname url@samestyle\endcsname
\providecommand{\newblock}{\relax}
\providecommand{\bibinfo}[2]{#2}
\providecommand{\BIBentrySTDinterwordspacing}{\spaceskip=0pt\relax}
\providecommand{\BIBentryALTinterwordstretchfactor}{4}
\providecommand{\BIBentryALTinterwordspacing}{\spaceskip=\fontdimen2\font plus
\BIBentryALTinterwordstretchfactor\fontdimen3\font minus
  \fontdimen4\font\relax}
\providecommand{\BIBforeignlanguage}[2]{{%
\expandafter\ifx\csname l@#1\endcsname\relax
\typeout{** WARNING: IEEEtran.bst: No hyphenation pattern has been}%
\typeout{** loaded for the language `#1'. Using the pattern for}%
\typeout{** the default language instead.}%
\else
\language=\csname l@#1\endcsname
\fi
#2}}
\providecommand{\BIBdecl}{\relax}
\BIBdecl

\bibitem{betts1998survey}
J.~T. Betts, ``Survey of numerical methods for trajectory optimization,''
  \emph{Journal of guidance, control, and dynamics}, vol.~21, no.~2, pp.
  193--207, 1998.

\bibitem{posa2014direct}
M.~Posa, C.~Cantu, and R.~Tedrake, ``A direct method for trajectory
  optimization of rigid bodies through contact,'' \emph{The International
  Journal of Robotics Research}, vol.~33, no.~1, pp. 69--81, 2014.

\bibitem{mordatch2012discovery}
I.~Mordatch, E.~Todorov, and Z.~Popovi{\'c}, ``Discovery of complex behaviors
  through contact-invariant optimization,'' \emph{ACM Transactions on Graphics
  (TOG)}, vol.~31, no.~4, p.~43, 2012.

\bibitem{tassa2012synthesis}
Y.~Tassa, T.~Erez, and E.~Todorov, ``Synthesis and stabilization of complex
  behaviors through online trajectory optimization,'' in \emph{Intelligent
  Robots and Systems (IROS), 2012 IEEE/RSJ International Conference on}.\hskip
  1em plus 0.5em minus 0.4em\relax IEEE, 2012, pp. 4906--4913.

\bibitem{tassa2014control}
Y.~Tassa, N.~Mansard, and E.~Todorov, ``Control-limited differential dynamic
  programming,'' in \emph{Robotics and Automation (ICRA), 2014 IEEE
  International Conference on}.\hskip 1em plus 0.5em minus 0.4em\relax IEEE,
  2014, pp. 1168--1175.

\bibitem{koenemann2015whole}
J.~Koenemann, A.~Del~Prete, Y.~Tassa, E.~Todorov, O.~Stasse, M.~Bennewitz, and
  N.~Mansard, ``Whole-body model-predictive control applied to the hrp-2
  humanoid,'' in \emph{Intelligent Robots and Systems (IROS), 2015 IEEE/RSJ
  International Conference on}.\hskip 1em plus 0.5em minus 0.4em\relax IEEE,
  2015, pp. 3346--3351.

\bibitem{mastalli}
C.~Mastalli, I.~Havoutis, M.~Focchi, D.~G. Caldwell, and C.~Semini,
  ``Hierarchical planning of dynamic movements without scheduled contact
  sequences,'' in \emph{Robotics and Automation (ICRA), 2016 IEEE International
  Conference on}.\hskip 1em plus 0.5em minus 0.4em\relax IEEE, 2016.

\bibitem{gehring2016}
C.~Gehring, S.~Coros, M.~Hutter, C.~D. Bellicoso, H.~Heijnen, R.~Diethelm,
  M.~Bloesch, P.~Fankhauser, J.~Hwangbo, M.~A. Hoepflinger, and R.~Y. Siegwart,
  ``{P}ractice {M}akes {P}erfect: {A}n {O}ptimization-{B}ased {A}pproach to
  {C}ontrolling {A}gile {M}otions for a {Q}uadruped {R}obot,'' \emph{IEEE
  Robotics \& Automation Magazine}, vol.~23, no.~1, pp. 34--43, 2016.

\bibitem{slq}
A.~Sideris and J.~E. Bobrow, ``An efficient sequential linear quadratic
  algorithm for solving nonlinear optimal control problems,'' \emph{Automatic
  Control, IEEE Transactions on}, vol.~50, no.~12, pp. 2043--2047, 2005.

\bibitem{hyq}
C.~Semini, N.~G. Tsagarakis, E.~Guglielmino, M.~Focchi, F.~Cannella, and D.~G.
  Caldwell, ``Design of hyq--a hydraulically and electrically actuated
  quadruped robot,'' \emph{Proceedings of the Institution of Mechanical
  Engineers, Part I: Journal of Systems and Control Engineering}, p.
  0959651811402275, 2011.

\bibitem{ilqg}
E.~Todorov and W.~Li, ``A generalized iterative lqg method for locally-optimal
  feedback control of constrained nonlinear stochastic systems,'' in
  \emph{Proceedings of the 2005, American Control Conference, 2005.}\hskip 1em
  plus 0.5em minus 0.4em\relax IEEE, 2005, pp. 300--306.

\bibitem{bloesch2013state}
M.~Bloesch, M.~Hutter, S.~Leutenegger, C.~Gehring, C.~D. Remy, and R.~Siegwart,
  ``State estimation for legged robots {--} consistent fusion of leg kinematics
  and {IMU},'' \emph{Robotics}, vol.~17, pp. 17--24, 2013.

\bibitem{mpc_quad}
M.~Neunert, C.~de~Crousaz, F.~Furrer, M.~Kamel, F.~Farshidian, R.~Siegwart, and
  J.~Buchli, ``Fast nonlinear model predictive control for unified trajectory
  optimization and tracking,'' in \emph{IEEE International Conference on
  Robotics and Automation (ICRA)}, 2016.

\bibitem{pratt2001virtual}
J.~Pratt, C.-M. Chew, A.~Torres, P.~Dilworth, and G.~Pratt, ``Virtual model
  control: An intuitive approach for bipedal locomotion,'' \emph{The
  International Journal of Robotics Research}, vol.~20, no.~2, pp. 129--143,
  2001.

\bibitem{ileg2015}
\BIBentryALTinterwordspacing
F.~Farshidian and J.~Buchli, ``Risk sensitive, nonlinear optimal control:
  Iterative linear exponential-quadratic optimal control with gaussian noise,''
  12 2015, arXiv, 1512.07173. [Online]. Available:
  \url{http://arxiv.org/abs/1512.07173}
\BIBentrySTDinterwordspacing

\bibitem{robcogen}
M.~Frigerio, J.~Buchli, D.~G. Caldwell, and C.~Semini, ``{R}ob{C}o{G}en: a code
  generator for efficient kinematics and dynamics of articulated robots, based
  on {D}omain {S}pecific {L}anguages,'' \emph{Journal of Software Engineering
  for Robotics (JOSER)}, 2016, accepted for publication.

\bibitem{victor_trot}
V.~Barasuol, J.~Buchli, C.~Semini, M.~Frigerio, E.~R. De~Pieri, and D.~G.
  Caldwell, ``A reactive controller framework for quadrupedal locomotion on
  challenging terrain,'' in \emph{Robotics and Automation (ICRA), 2013 IEEE
  International Conference on}.\hskip 1em plus 0.5em minus 0.4em\relax IEEE,
  2013, pp. 2554--2561.

\bibitem{coros2011locomotion}
S.~Coros, A.~Karpathy, B.~Jones, L.~Reveret, and M.~Van De~Panne, ``Locomotion
  skills for simulated quadrupeds,'' in \emph{ACM Transactions on Graphics
  (TOG)}, vol.~30, no.~4.\hskip 1em plus 0.5em minus 0.4em\relax ACM, 2011,
  p.~59.

\end{thebibliography}
